\documentclass[10pt,journal,compsoc]{IEEEtran}

\usepackage{amsmath}
\usepackage{amssymb}  
\usepackage{booktabs}
\usepackage{array}
\usepackage{pifont} 
\usepackage[table]{xcolor}
\definecolor{bestgray}{gray}{0.80} 
\definecolor{secondgray}{gray}{0.9}

\ifCLASSOPTIONcompsoc
  \usepackage[nocompress]{cite}
\else
  \usepackage{cite}
\fi
\ifCLASSINFOpdf
\else
\fi
\usepackage{amsmath}
\usepackage{graphicx}

\usepackage{xcolor}

\makeatletter
\@namedef{cb@soft}{}
\@namedef{cb@liu2025channel}{}
\@namedef{cb@zeng2021toward}{}
\@namedef{cb@zeng2024tutorial}{}
\@namedef{cb@li2022channel}{}
\@namedef{cb@nukapotula2025gsparc}{}

\begin{document}
%
\title{Deformable 2D Gaussian Splatting for Efficient Wireless Radiance Field Rendering}
%
%
%
%

\author{Mufan Liu*, Cixiao Zhang*, Qi Yang, \IEEEmembership{Member,~IEEE,} Yujie Cao, Yiling Xu${}^\dagger$, \IEEEmembership{Member,~IEEE,}
\\Yin Xu${}^\dagger$, \IEEEmembership{Senior Member,~IEEE,}  Shu Sun, \IEEEmembership{Senior Member,~IEEE}, Mingzeng Dai, Yunfeng Guan
\IEEEcompsocitemizethanks{\IEEEcompsocthanksitem M. Liu, C. Zhang, Y. Cao, Y. Xu, Y. Xu, S. Sun and Y. Guan are from School of Information Science and Electronic Engineering, Shanghai Jiao Tong University, Shanghai,
200240, China (email: \{sudo\_evan, cixiaozhang, 1452377525, yl.xu, xuyin shusun, yfguan69\}@sjtu.edu.cn)
\IEEEcompsocthanksitem Q. Yang is with University of Missouri-Kansas City, Kansas, 64110, America
(email: qiyang@umkc.edu)

\IEEEcompsocthanksitem M. Dai is with  Lenovo Group, Lenovo Research, Beijing, 2100094, China
(email: \{daimz4\}@lenovo.com)}

\thanks{(*Authors contribute equally to this work.)\\(${}^\dagger$Corresponding authors.)}}

%
%

\markboth{Journal of \LaTeX\ Class Files,~Vol.~14, No.~8, August~2015}%
{Shell \MakeLowercase{\textit{et al.}}: Bare Advanced Demo of IEEEtran.cls for IEEE Computer Society Journals}
%



\IEEEtitleabstractindextext{%
\begin{abstract}

Modeling the wireless radiance field (WRF) is fundamental to modern communication systems, enabling key tasks such as localization, sensing, and channel estimation. Traditional approaches, which rely on empirical formulas or physical simulations, often suffer from limited accuracy or require strong scene priors. Recent neural radiance field (NeRF)–based methods improve reconstruction fidelity through differentiable volumetric rendering, but their reliance on computationally expensive multilayer perceptron (MLP) queries hinders real-time deployment. To overcome these challenges, we introduce Gaussian splatting (GS) to the wireless domain, leveraging its efficiency in modeling optical radiance fields to enable compact and accurate WRF reconstruction. Specifically, we propose SwiftWRF, a deformable 2D Gaussian splatting framework that synthesizes WRF spectra at arbitrary positions under single-sided transceiver mobility. SwiftWRF employs CUDA-accelerated rasterization to render spectra at over 100k FPS and uses the lightweight MLP to model the deformation of 2D Gaussians, effectively capturing mobility-induced WRF variations. In addition to novel spectrum synthesis, the efficacy of SwiftWRF is further underscored in its applications in angle-of-arrival (AoA) and received signal strength indicator (RSSI) prediction. Experiments conducted on both real-world and synthetic indoor scenes demonstrate that SwiftWRF can reconstruct WRF spectra up to 500x faster than existing state-of-the-art methods, while significantly enhancing its signal quality. The project page is https://evan-sudo.github.io/swiftwrf/.


\end{abstract}

}

\maketitle

\IEEEdisplaynontitleabstractindextext

%
\IEEEpeerreviewmaketitle

\ifCLASSOPTIONcompsoc
\IEEEraisesectionheading{\section{Introduction}\label{sec:introduction}}
\else

\section{Introduction}
\label{sec:introduction}
\fi

\begin{figure}[t]
    \centering
    \includegraphics[width=\linewidth]{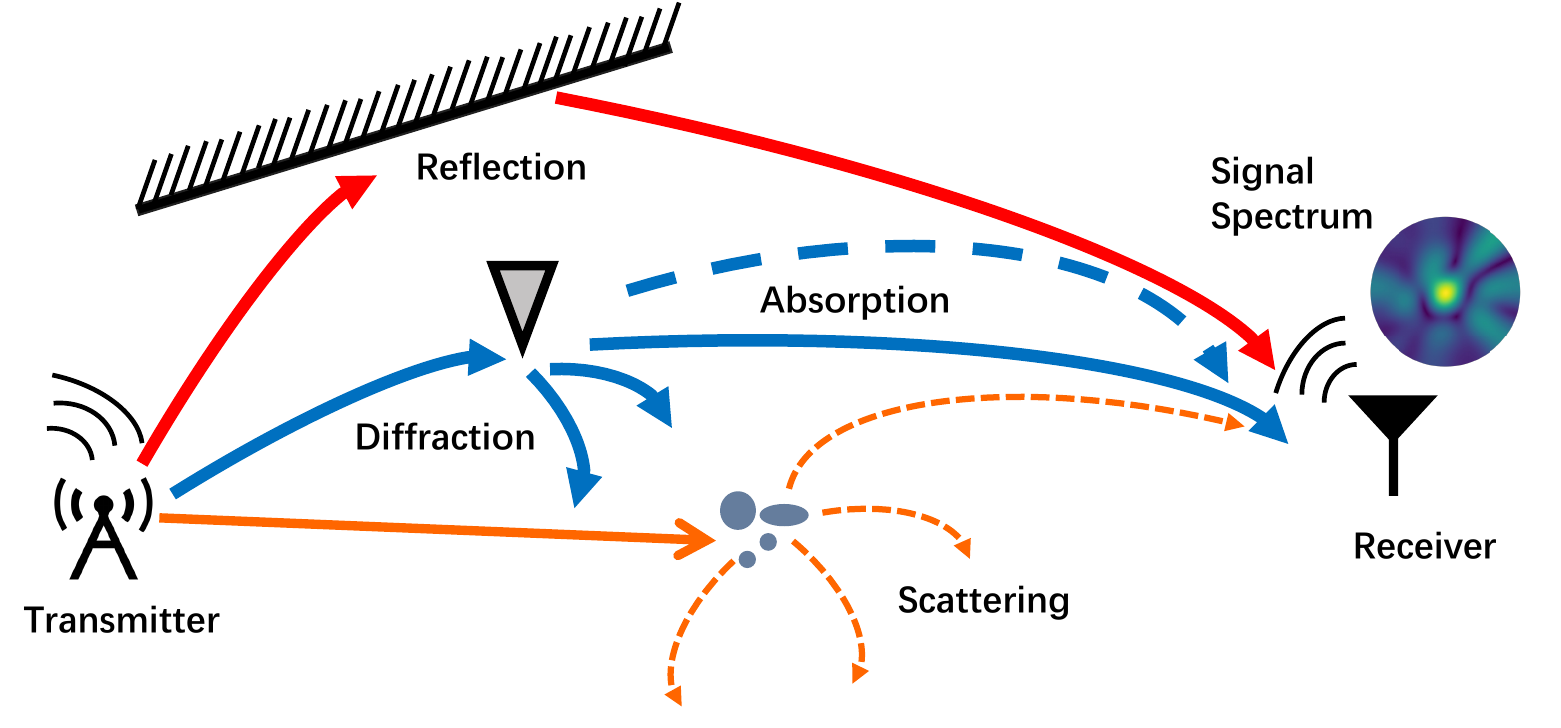}
    \caption{A toy example illustrating the wireless radiance field from a given TX-RX deployment.}
    \label{toyexample}
    \vspace{-0.6cm}
\end{figure}

In modern communication systems, digital information is conveyed through radio frequency (RF) waveforms that interact with the physical environment. As these signals propagate from a transmitter (TX) to a receiver (RX), they undergo reflection, diffraction, scattering, and absorption, giving rise to a continuous electromagnetic signal field (see Fig. \ref{toyexample}). To understand and utilize this complex propagation behavior, Wireless Radiance Field (WRF) modeling has emerged as a means to learn how the signal field's frequency response varies across space \cite{DeAlwis2021Survey}. The task is to reconstruct the spatial spectrum, which captures the omnidirectional signal distribution observed at an RX, along novel TX–RX deployments, using spectra collected from known ones. WRF modeling encapsulates rich geometric and spectral information that benefits downstream tasks such as channel estimation, localization, and environmental sensing. Over the years, researchers have explored a wide range of techniques to tackle this problem, from classical Maxwell-based formulations to modern deep neural network approaches. As 6G technologies continue to advance, communication systems in latency-critical domains such as the Internet of Things (IoT) are expected to deliver ultra-reliable, low-latency performance, with end-to-end delays ranging from 10 to 100 microseconds \cite{6GIOTsurvey, 9598915, zeng2021toward}. This makes fast and accurate WRF modeling essential for the practical deployment of next-generation wireless systems.

\begin{figure}[t]
    \centering
    \includegraphics[width=0.995\linewidth]{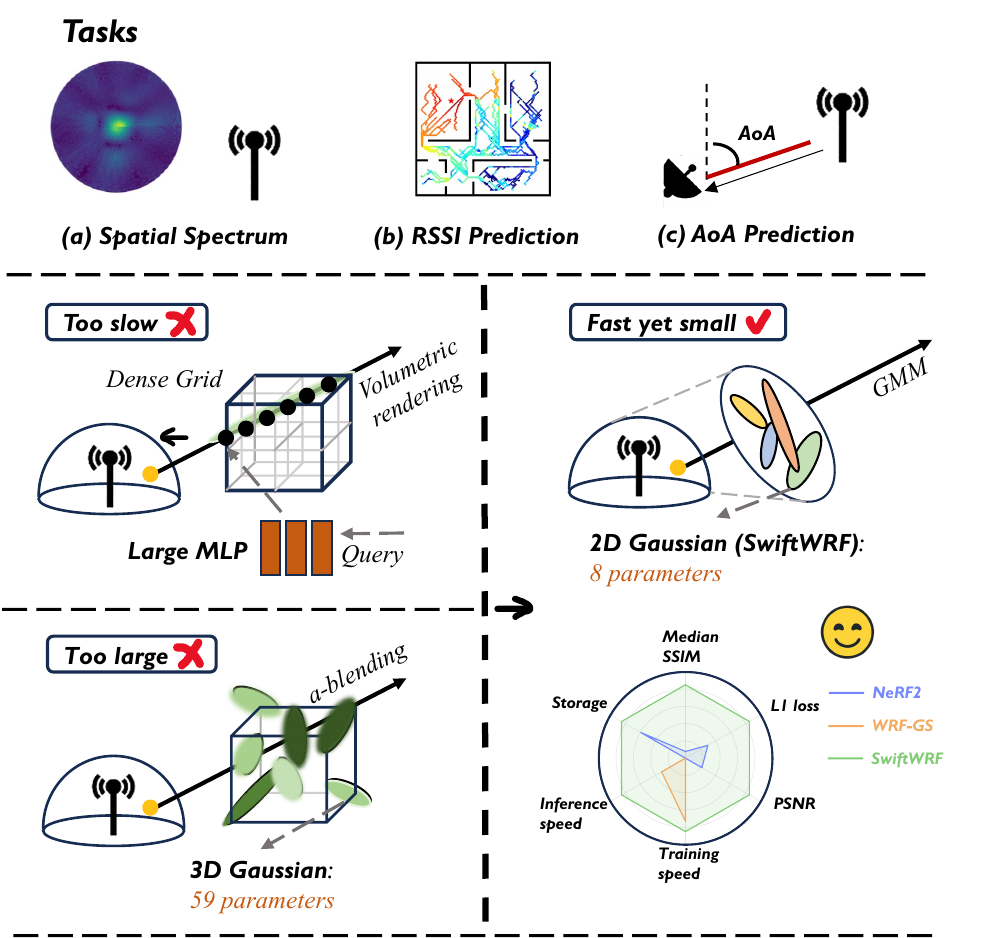}
    \caption{\textbf{Comparison with NeRF-based and GS-based WRF modeling approaches.} NeRF-based solutions rely on computationally intensive volumetric rendering, while 3DGS suffers from dimensional redundancy due to an invariant projection. In contrast, our method achieves both efficiency and compactness by adopting a streamlined 2D Gaussian formulation tailored to the wireless domain.}
    \label{motivation}
\end{figure}

\textbf{Background:}
WRF modeling aims to represent how spatial spectra vary with transceiver location \cite{liu2025channel, zeng2024tutorial}. Traditional rule-based approaches, including empirical path-loss models and ray tracing, offer a speed-accuracy trade-off: empirical models are fast but often miss fine-grained spatial variations, while ray tracing can be accurate but depends on detailed scene models and is computationally expensive \cite{wrf,rulebasedwrf1,rulebasedwrf2,schmitz2010efficient,he2018design}. 
Learning-based methods improve fidelity by leveraging deep generative models (e.g., variational autoencoders (VAEs) and generative adversarial networks (GANs)) to synthesize unobserved spectra from measurements \cite{vae,gan}. A recent line of work draws a direct analogy between WRF reconstruction and novel view synthesis (NVS), treating spectrum prediction at unseen locations as rendering from a neural radiance field (NeRF) \cite{yuan2025sed, Nerf2}. NeRF-style wireless methods adopt multilayer perceptron (MLP)-based volumetric rendering and achieve strong quality \cite{mildenhall2021nerf,Nerf2,newrf,chen2024rfcanvas}, yet their inference remains expensive because each query requires many MLP evaluations along rays (or dense samples) and the cost grows with the target resolution and the number of queried locations. This limits practicality in latency-sensitive settings.

Gaussian splatting (GS) offers a more efficient rendering paradigm by representing the field with explicit Gaussian primitives and computing outputs via CUDA-parallel rasterization \cite{kerbl20233d,3dgstvcg}. 
Compared with volumetric rendering, GS avoids repeated network queries at inference and instead aggregates contributions from a set of differentiable primitives in a tile-wise manner, which is well suited to real-time synthesis. These properties motivate us to adopt a GS-style explicit representation for WRF, so that spectra at novel positions can be reconstructed with high throughput while preserving reconstruction quality.


\textbf{Observation: }While GS represents the state-of-the-art (SOTA) in field reconstruction, directly applying it to WRF modeling exposes several mismatches between the assumptions of the vision domain and the wireless domain. These gaps hinder the efficiency and effectiveness of GS in this context. \textit{First, 3D volumetric representation is redundant in the WRF setting.} The WRF modeling task is typically defined as synthesizing spatial spectra at a fixed RX, where a moving TX emits distinct WRFs from different positions. In this setting, each WRF is modeled using a separate set of 3D Gaussians, followed by a constant mapping to form the RX’s spectrum. Since the RX’s position and orientation remain unchanged, only the components of the 3D Gaussians aligned with its view contribute to the observed signal (see Fig. \ref{motivation} in the middle). This causes the 3D representation to degenerate into 2D Gaussians, rendering the volumetric design unnecessary. \textit{Second, existing methods lack locality-aware modeling.} In \cite{wrf}, 3D Gaussian attributes are predicted independently at each TX location. However, as the TX follows a continuous trajectory and wireless propagation is governed by deterministic multipath effects, the resulting WRF varies smoothly in space. This implies strong correlations between spectra observed at neighboring TX locations. \textit{Third, sparse supervision introduces a risk of overfitting.} Existing methods optimize quality loss only over observed regions, without applying any regularization to unobserved areas. This lack of supervision hinders the model’s ability to interpolate and generalize to unseen positions. \textit{Fourth, limited evaluation settings.} Existing methods are typically tested on a single scene under the TX-side moving configuration ($\text{NeRF}^2-\text{s}23$ dataset \cite{Nerf2}). The lack of diverse scenes, i.e., RX-side moving cases, limits the credibility of their results. Together, these limitations constrain the scalability, generalization, and physical alignment of existing GS-based WRF modeling approaches. In response, we propose a streamlined GS framework tailored to the structural characteristics of the WRF, and enabling evaluation across both TX-side and RX-side moving scenarios.

\textbf{Our method: }We formulate the WRF modeling problem under a single-sided transceiver mobility setting, where the perceived WRF spectra between the TX–RX pair vary due to the motion of one moving side. This setting follows the evaluation protocol of NeRF$^2$ and related WRF surrogates, but we depart from their MLP-based implicit field representation: instead of querying an MLP densely for rendering, we extend the explicit rasterization line of Gaussian splatting to the wireless spectrum domain. In particular, our representation inherits the 2D Gaussian rasterization idea used in image/video 2DGS works \cite{zhang2024gaussianimage, liu2025d2gv}, and adapts it to model antenna-array spectrum outcomes under mobility. To address this, we propose a deformable 2D Gaussian splatting (2DGS) framework, termed SwiftWRF. The method begins by formulating a 2D Gaussian representation tailored to the wireless domain, where each primitive captures localized signal behavior through a 2D Gaussian distribution and a complex-valued response. Unlike prior image/video 2DGS methods defined on the image plane, our primitives are formulated in the polar spectrum domain for antenna-array rasterization. We adopt a Cholesky-parameterized covariance and replace RGB appearance modeling with signal attenuation and a complex-valued response. To render the spatial spectrum, we adopt a Gaussian mixture model (GMM)–like formulation~\cite{zhang2024gaussianimage}, in which all 2D Gaussian primitives are directly blended over the plane via differentiable rasterization. Compared to 3D Gaussians, our 2D formulation substantially improves compactness, reducing the parameter count by a factor of seven, as illustrated in Fig.~\ref{motivation}. The blending process also eliminates the need for camera projection and removes dependence on geometry-based initialization. Complementing this, transceiver mobility is modeled through a continuous deformation field, implemented as a lightweight MLP that predicts position-dependent offsets to a global set of 2D Gaussian parameters. This design enables fast and spatially coherent adaptation across locations, effectively translating the spatial deformation of the WRF into a planar deformation in the 2D Gaussian space. For model training, we adopt a coarse-to-fine training protocol: the model is first trained without deformation on a shared set of 2D Gaussians, and subsequently fine-tuned with the deformation field enabled. To further improve generalization and mitigate overfitting, we introduce an annealed smoothing mechanism that injects controlled positional noise during fine-tuning. The noise magnitude decays exponentially over time, promoting smoother optimization in the early stages and higher reconstruction quality in the later stages.

For evaluation, we construct a synthetic dataset to assess the generality of SwiftWRF beyond TX-side moving settings. The dataset includes both TX-side moving and RX-side moving scenarios, spanning three representative indoor layouts, with spectra generated using MATLAB’s ray-tracing toolbox. Extensive experiments show that SwiftWRF achieves a peak signal-to-noise ratio (PSNR) improvement of 2.33--10.41~dB over existing baselines, while delivering inference speeds of approximately 100k FPS—over 500$\times$ faster than WRF-GS and more than 20{,}000$\times$ faster than NeRF$^2$. Finally, we evaluate SwiftWRF on two case studies. For angle-of-arrival (AoA) prediction, the synthesized spectra are integrated into a turbo-learning framework, where they serve as augmented data to enhance prediction accuracy. For received signal strength indicator (RSSI) prediction, we adapt the rendering head with average pooling to support scalar field regression. These case studies validate the generalizability of our method and demonstrate consistent improvements over existing baselines.\\
\noindent \textbf{To summarize, our contributions are as follows:}
\begin{itemize}
    \item We propose a compact and efficient 2DGS representation tailored to WRF modeling, which replaces the unnecessary 3D structure and spherical harmonics used in prior methods.
    
    \item We introduce a deformation field module parameterized by a lightweight MLP to effectively capture transceiver-induced variations across space, along with an annealed smoothing strategy to enhance generalization.

    \item We construct and release a synthetic WRF dataset covering both TX-side moving and RX-side moving scenarios, generated using MATLAB's ray-tracing toolbox over multiple representative indoor environments.

    \item We extensively evaluate our method on WRF reconstruction and two downstream tasks, AoA and RSSI prediction, and demonstrate significant gains in both accuracy and efficiency.
\end{itemize}
The remainder of this paper is organized as follows. Section~2 reviews related work on GS and wireless radiance fields. Section~3 introduces necessary preliminaries and formalizes the problem statement. Section~4 details the proposed SwiftWRF framework for WRF modeling under single-sided mobility. Section~5 describes the construction of our synthetic spectrum dataset. Section~6 presents the experimental setup and evaluation results. Section~7 showcases two case studies, and Section~8 concludes the paper and outlines future research directions.

\section{Related Work}
Our work falls within the domain of WRF modeling and channel parameter prediction, and intersects with the emerging area of GS.
\subsection{Wireless Radiance Field Modeling}
WRF modeling seeks to characterize electromagnetic wave propagation between TXs and RXs, incorporating reflection, diffraction, scattering, and path‐loss effects \cite{oestges2002deterministic, tong2018cooperative}. Conventional approaches to channel modeling can be broadly classified into two categories: probabilistic models and deterministic ray-tracing models. The former are computationally efficient and easy to implement; however, they often fail to capture fine-grained spatial variations and site-specific details \cite{raytracing, li2022channel}. The latter offer high-fidelity, site-specific predictions by modeling the exact propagation environment \cite{DeAlwis2021Survey}, but they are computationally intensive and require detailed prior knowledge of the geometric layout, which limits their scalability and applicability in dynamic or unknown environments. Learning‐based methods instead train deep networks on extensive, location‐tagged channel measurements to directly infer complex propagation phenomena \cite{vae, gan}, but they often lack interpretability and struggle to generalize beyond their training distributions. More recently, NeRF--based methods have been applied to learn WRF as an optical field \cite{Nerf2, newrf, orekondy2023winert, amballa2025can}. Zhao et al.~\cite{Nerf2} developed $\text{NeRF}^2$ for spatial-spectrum reconstruction in real environments; Orekondy et al.~\cite{orekondy2023winert} introduced a NeRF surrogate for indoor EM propagation using synthetic datasets; Lu et al.~\cite{newrf} proposed NeWRF, which predicts channel states from sparse measurement sets; and EchoNeRF \cite{amballa2025can} learns a NeRF-style RF field from sparse RF measurements to infer indoor structure by modeling the received signal power as a combination of LoS and multipath reflection components. Considering the slow rendering of NeRF, several GS-based methods model WRF with explicit Gaussian primitives and leverage CUDA-accelerated rasterization for near real-time synthesis. RF-3DGS \cite{zhang2024rf} reconstructs a radio radiance field and targets multimodal spatial-channel state information (CSI), e.g., gain/delay/AoA, by encoding channel attributes with CSI-aware spherical-harmonics representations, enabling fast CSI queries at arbitrary locations. RFSPM \cite{yang2025scalable} focuses on scalable RF signal spatial propagation modeling, representing a scene with customized GPU kernels to efficiently predict received signal characteristics across diverse wireless settings. More recently, GSpaRC \cite{nukapotula2025gsparc} pushes GS-based RF reconstruction toward real-time deployment by achieving sub-millisecond latency for RF channel/CSI reconstruction while maintaining competitive accuracy. However, these approaches remain rudimentary 3D GS frameworks and lack the specialized design needed to meet the stringent requirements of ultra-reliable low-latency communication (URLLC) \cite{liu2025channel, soft}. To address this gap, we present SwiftWRF, an efficient 2DGS framework that simultaneously achieves faster training and inference times, alongside superior reconstruction fidelity.

\subsection{Channel Parameter Prediction}
\textbf{RSSI prediction:} RSSI prediction estimates signal strength across a venue from limited measurements, underpinning coverage planning and indoor localization. Shin et al. \cite{mri} proposed MRI, a smartphone‐based system that interpolates indoor RSSI via a log‐distance path‐loss model and sparse war‐walking data. Zhao et al. \cite{Nerf2} developed a $\text{NeRF}^2$‐based, turbo‐learning RSSI predictor using the public BLE dataset \cite{Nerf2}. We similarly retrain SwiftWRF on that dataset to evaluate its RSSI prediction performance. \\
\textbf{AoA prediction:} AoA prediction estimates the direction of line-of-sight (LOS) propagation by identifying the peak in the spatial spectrum, serving as a critical component for accurate indoor localization. However, the task remains highly challenging due to multipath propagation and destructive interference of varying wireless conditions. To address this issue, iArk’s AoA Neural Network (AANN) \cite{iArk} corrects spectral peak deviations via deep learning but requires large volumes of high-quality data. $\text{NeRF}^2$’s turbo-learning strategy \cite{Nerf2} alleviates this by synthesizing spatial spectra to augment the training set. Building on these, we apply the turbo-learning strategy to the synthesized dataset generated by SwiftWRF and retrain AANN on a mixed dataset generated by SwiftWRF, which yields cleaner spectra and further boosts AoA prediction in complex wireless scenarios.

\subsection{Gaussian Splatting}
GS has recently emerged as a powerful technique for novel‐view synthesis, employing explicit 3D Gaussian primitives and differentiable, tile‐based rasterization to achieve real‐time rendering and interactive editing capabilities \cite{kerbl20233d, sun2024splatter}. Its efficacy has motivated extensions to dynamic scene modeling \cite{yang2024deformable, wu20244d}, AI‐driven content creation \cite{tang2023dreamgaussian}, autonomous driving \cite{zhou2024drivinggaussian}, and wireless communication \cite{wrf, yang2025scalable, li2025wideband}. Although initially developed for 3D applications, the principles of GS have also been extended to 2D domains \cite{liu2025d2gv}. For example, GaussianImage \cite{zhang2024gaussianimage} represents images using 2D Gaussian kernels to enable rapid decoding, while Hu et al. \cite{huang20242d} employ 2D Gaussians to model precise mesh geometry. To our knowledge, SwiftWRF is the first framework to leverage 2DGS for efficient WRF modeling. Across multiple indoor scenarios, SwiftWRF delivers state‐of‐the‐art reconstruction accuracy and furthermore facilitates downstream RF tasks such as AoA and RSSI prediction.

\begin{figure}[t]
    \centering
    \includegraphics[width=0.99\linewidth]{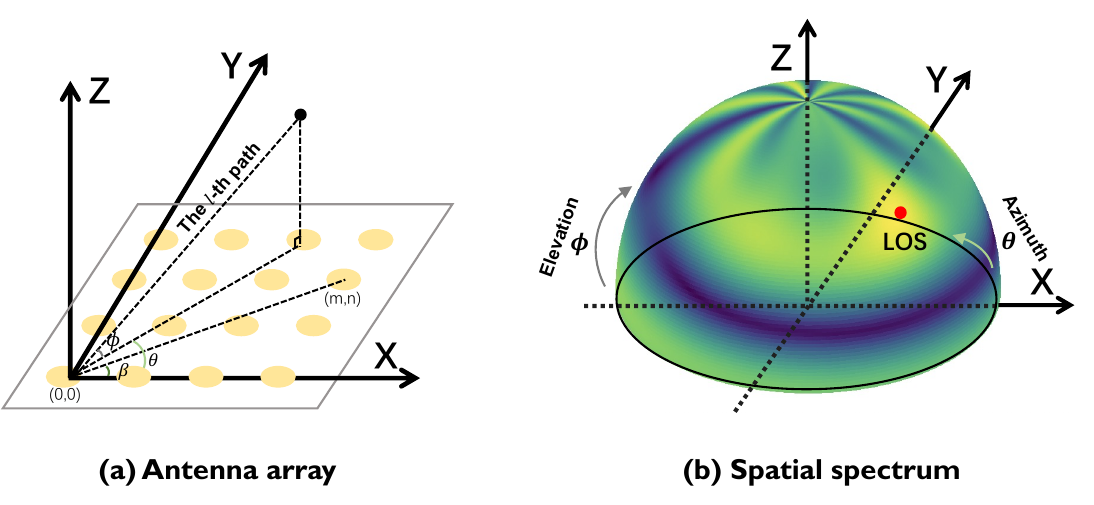}
    \caption{Illustration of RX (origin) antenna array (a) and spatial spectrum (b).}
    \label{fig:spatial-spectrum}
\end{figure}

\section{Preliminary and Problem Formulation}
In this section, we first review the core principles of WRF and formalize the WRF modeling problem under single-sided transceiver mobility. We then introduce the key concepts of GS that are adapted to the WRF domain.
\subsection{Wireless Radiance Field}

A typical wireless communication link involves a TX and an RX that exchange information via RF signals. The TX emits an RF waveform that propagates through space and eventually reaches the RX. During propagation, the signal interacts with the environment through reflection, diffraction, scattering, and absorption. These interactions give rise to multiple propagation paths, whose superposition at different spatial locations forms a spatially varying and complex electromagnetic field—known as the WRF.

To extract directional information from this field, the RX is typically equipped with an antenna array composed of spatially distributed elements. By adjusting the phase or timing of the signals received at each element, the array can steer its sensitivity toward a desired direction, effectively forming a narrow reception beam. Sweeping across angles allows the RX to measure signal energy from multiple directions and aggregate these measurements into a spatial spectrum, as illustrated in Fig.~\ref{fig:spatial-spectrum}.

\textbf{Spatial spectrum:} Consider an RX equipped with a two-dimensional \(\sqrt{K} \times \sqrt{K}\) antenna array with element spacing \(D\), as illustrated in Fig.~\ref{fig:spatial-spectrum}. The \((m, n)\)-th antenna element, denoted \(R_{m,n}\), is located at a position specified by its polar coordinates. Its radial and angular coordinates are defined as
\begin{equation}
    \left\{
    \begin{aligned}
    r_{m,n} &= D \sqrt{(m-1)^2 + (n-1)^2}, \\
    \beta_{m,n} &= \arctan2(m-1, n-1).
    \end{aligned}
    \right.
\end{equation}
Assuming $L_p$ multipath components, the narrowband channel at $R_{m,n}$ is
\begin{equation}
h_{m,n} = \sum_{l=1}^{L_p} \rho_l e^{j\Delta\alpha_{m,n}(\hat{\theta}_l,\hat{\phi}_l)} , \label{eq:received signal}
\end{equation}
where $\rho_l$, $\hat{\theta}_l$, and $\hat{\phi}_l$ are the power coefficient, azimuth angle, and elevation angle, respectively, all associated with the $l$-th propagation path. According to \cite{Nerf2}, the phase difference is expressed as
\begin{align}
\Delta \alpha_{m,n}(\hat{\theta}_l,\hat{\phi}_l) &= \alpha_{m,n}(\hat{\theta}_l,\hat{\phi}_l) - \alpha_{1,1}(\hat{\theta}_l,\hat{\phi}_l) \nonumber\\
&= 2\pi \Delta d_{m,n}(\hat{\theta}_l,\hat{\phi}_l) / \lambda \mod 2\pi ,
\label{eq:phase shift}
\end{align}
where $\Delta d_{m,n}(\hat{\theta}_l,\hat{\phi}_l)=-r_{m,n} \cos(\hat{\theta}_l - \beta_{m,n}) \cos(\hat{\phi}_l)
$ represents the path length difference between antennas $R_{m,n}$ and the reference antenna $R_{1,1}$ along direction $(\hat{\theta}_l,\hat{\phi}_l)$, and $\lambda$ denotes the signal wavelength. The RX estimates the spatial spectrum by applying an analog combining matrix that steers beams toward predefined azimuth $\theta$ and elevation $\phi$ angles. When a beam aligns with the true AoA, signals from that direction combine constructively, while signals from other directions interfere destructively and are attenuated~\cite{iArk}. According to~\cite{Nerf2,wrf}, the spatial spectrum at direction $(\theta,\phi)$ is then computed as
\begin{equation}
\mathbf{A}(\theta, \phi)
= \left\lvert
\frac{1}{K}
\sum_{m=1}^{\sqrt{K}}
\sum_{n=1}^{\sqrt{K}}
w_{m,n}(\theta,\phi)\,e^{j\angle h_{m,n}}
\right\rvert,
\label{eq:beam}
\end{equation}
where $w_{m,n}(\theta,\phi)=e^{-j\Delta\alpha_{m,n}(\theta,\phi)}$ is the complex weight for steering a beam towards the angle $(\theta, \phi)$, applied to the $(m,n)$-th antenna element, and $\angle h_{m,n}$ denotes the phase of $h_{m,n}$. Under our assumption of \(1^\circ\) angular resolution, the spatial spectrum is represented as a \(360 \times 90\) matrix \(\mathbf{A}\), defined by
\begin{equation}
\mathbf{A} =
\begin{bmatrix}
\mathbf{A}(0^\circ,0^\circ) & \mathbf{A}(1^\circ,0^\circ) & \cdots & \mathbf{A}(359^\circ,0^\circ) \\
\mathbf{A}(0^\circ,1^\circ) & \mathbf{A}(1^\circ,1^\circ) & \cdots & \mathbf{A}(359^\circ,1^\circ) \\
\vdots & \vdots & \ddots & \vdots \\
\mathbf{A}(0^\circ,89^\circ) & \mathbf{A}(1^\circ,89^\circ) & \cdots & \mathbf{A}(359^\circ,89^\circ)
\end{bmatrix}.
\end{equation}
The spatial spectrum represents a spatial power distribution function that directly captures the propagation behavior of RF signals through the environment. As such, it is commonly used as an effective representation of the WRF.

\subsection{Gaussian Splatting}
GS originally refers to the 3DGS framework, which performs NVS through CUDA-based differentiable rasterization. Beyond its success on NVS benchmarks, its 2D extension (2DGS) enables more compact representations of 2D surfaces, including images, meshes, and point-based geometry. Since the WRF is received in the form of a 2D spatial spectrum, it is naturally constructed using a 2D Gaussian-based formulation for improved efficiency. To motivate this choice, the following section first reviews the 3DGS framework, followed by a description of the 2DGS.\\
\textbf{3D Gaussian: }3DGS reconstructs unseen views of a 3D scene by training a set of Gaussian primitives from a sparse collection of captured images \cite{kerbl20233d}. These primitives are positioned in free 3D space, with each representing a small, smooth region characterized by a set of learnable attributes: a mean position \(\boldsymbol{\mu} \in \mathbb{R}^3\), a covariance matrix \(\boldsymbol{\Sigma} \in \mathbb{R}^{3 \times 3}\), an opacity value \(\alpha \in [0,1]\) indicating transparency, and a color vector \(\boldsymbol{c} \in \mathbb{R}^k\) encoding appearance via spherical harmonics (SH) coefficients. Each Gaussian primitive models the likelihood of a point \(\boldsymbol{x} \in \mathbb{R}^3\) belonging to its region, following a 3D Gaussian distribution defined as
\begin{equation}
\mathbf{G}(\boldsymbol{x}) = \exp\left( -\frac{1}{2} (\boldsymbol{x} - \boldsymbol{\mu})^\top \boldsymbol{\Sigma}^{-1} (\boldsymbol{x} - \boldsymbol{\mu}) \right).
\end{equation}
For rendering, primitives are sorted by depth in camera space and projected through the camera matrix onto the image plane, producing 2D Gaussian footprints. These footprints are blended front to back: closer primitives contribute first, while more distant ones are attenuated by the accumulated opacity. At each pixel, the color $C$ is obtained by $\alpha$-blending all overlapping footprints along the viewing ray
\begin{equation}
C = \sum_{i=1}^{N} \boldsymbol{c}_i \alpha_i \prod_{j=1}^{i-1}(1-\alpha_j),
\end{equation}
where $\boldsymbol{c}_i$ and $\alpha_i$ denote the color and opacity of the $i$-th primitive, respectively, and $N$ is the number of primitives affecting the pixel. This process is known as differentiable rasterization, as it enables the optimization of Gaussian attributes through gradient backpropagation based on the end-to-end rendering loss. \\
\textbf{2D Gaussian:} While modeling the full WRF is inherently a 3D problem, synthesizing the RX’s spatial spectrum does not require 3D projection and can be captured more efficiently using 2D Gaussian primitives. In 2DGS, each Gaussian primitive is parameterized by a 2D position $\boldsymbol{\mu} \in \mathbb{R}^2$, a 2D covariance matrix $\boldsymbol{\Sigma} \in \mathbb{R}^{2\times 2}$, and a DC color vector $\boldsymbol{c} \in \mathbb{R}^3$. Higher-order spherical harmonics coefficients are discarded as the viewpoint is fixed in the 2D setting. To ensure the positive semi-definiteness of $\boldsymbol{\Sigma}$ and maintain numerical stability during optimization, a Cholesky factorization is employed, where the covariance matrix is decomposed as $\boldsymbol{\Sigma} = L L^\top$, with $L \in \mathbb{R}^{2\times 2}$ being a lower triangular matrix parameterized by $\boldsymbol{l} = \{l_1, l_2, l_3\}$. To render a 2D image, the color $C$ at each pixel is computed by aggregating the contributions from all nearby Gaussians $N$, following
\begin{equation}
C = \sum_{n=1}^{N} \boldsymbol{c}_n \cdot \exp\left( -\frac{1}{2} \boldsymbol{d}_n^\top \boldsymbol{\Sigma}_n^{-1} \boldsymbol{d}_n \right),
\end{equation}
where $\boldsymbol{d}_n = \boldsymbol{x}_i - \boldsymbol{\mu}_n$ denotes the displacement between the pixel location $\boldsymbol{x}_i$ and the center of the $n$-th Gaussian. 2D Gaussians were originally introduced for fast image/content reconstruction. In contrast, our goal is to predict wireless spectra at unseen positions, which requires both strategies to mitigate overfitting and a wireless-domain adaptation of the 2D Gaussian formulation to better capture signal-specific characteristics.


\subsection{Problem Formulation}
\subsubsection{Problem Statement}


We study the problem of reconstructing spatial spectra at unobserved transceiver positions using spectra collected from known locations, under a single-sided transceiver mobility setting. In this setting, either the TX or the RX moves in a multipath-rich environment, while both maintain fixed orientations. In the TX-side moving case, the WRF is dynamically emitted from the moving TX, while the mapping from the WRF to the observed spectrum remains constant due to a fixed RX. Given the constant field-to-spectrum mapping, the WRF variations are directly translated into changes in the measured spectrum. This reduces the problem to modeling how the spectrum changes with TX position, which can be viewed as a 2D interpolation task over space. In the RX-side moving case, the WRF is static, and the RX samples the spectrum from different positions. This corresponds to a 3D reconstruction problem, for which 3D Gaussian modeling is a natural fit. However, since the WRF itself does not change, the variation in the observed spectrum is entirely due to the change in the field-to-spectrum mapping, which is a function of the RX position. This perspective allows the task to be reformulated as a 2D interpolation problem over RX positions, consistent with the TX-side moving case. 


\subsubsection{Problem Definition}
We now formally define the task. To begin, we describe the two settings considered under single-sided transceiver mobility:

\medskip
\noindent\textit{TX-side moving scenario.} The TX moves across a set of known positions
\[
\mathcal{P} = \{\boldsymbol{p}_i \in \mathbb{R}^3\}_{i=1}^{N_s},
\]
and emits WRF signals at each location, while the RX remains fixed at \(\boldsymbol{q} \in \mathbb{R}^3\) to collect the corresponding spectra \(\{\mathbf{A}_i\}_{i=1}^{N_s}\). Here, \(N_s\) denotes the number of TX positions.

\medskip
\noindent\textit{RX-side moving scenario.} The TX emits WRF signals from a fixed position \(\boldsymbol{p} \in \mathbb{R}^3\), while the RX moves across a set of known positions
\[
\mathcal{Q} = \{\boldsymbol{q}_i \in \mathbb{R}^3\}_{i=1}^{N_s},
\]
and collects the corresponding spectra \(\{\mathbf{A}_i\}_{i=1}^{N_s}\). Again, \(N_s\) denotes the number of RX positions.\\
The indoor scene is assumed to exhibit rich multipath propagation. To prevent Doppler effects, the moving transceiver remains stationary during each spectrum acquisition. At each position \(\boldsymbol{s}_i \in \mathcal{P}\) (\(\mathcal{Q}\) for RX-moving), the spectrum is collected by RX as a discrete angular spectrum matrix
\[
\mathbf{A}_i = \bigl[\mathbf{A}_i(\theta_j, \phi_k)\bigr] \in \mathbb{C}^{M_\theta \times M_\phi},
\]
where \(M_\theta\) and \(M_\phi\) denote the number of angular samples in the azimuth and elevation directions, respectively. The angular grids are given by \(\{\theta_j\}_{j=1}^{M_\theta} \subset [0, 2\pi)\) and \(\{\phi_k\}_{k=1}^{M_\phi} \subset \left[0, \tfrac{\pi}{2}\right]\). Let
\[
\{(\boldsymbol{s}_i, \mathbf{A}_i)\}_{i=1}^{N_s},
\quad
\boldsymbol{s}_i \in 
\begin{cases}
\mathcal{P}, & \text{TX-side moving},\\
\mathcal{Q}, & \text{RX-side moving},
\end{cases}
\]
denote the collected spectra paired with their corresponding transceiver positions. The objective is to learn a continuous mapping
\[
\mathcal{F}: \mathbb{R}^3 \to \mathbb{C}^{M_\theta \times M_\phi}, \quad \boldsymbol{s} \mapsto \mathcal{F}(\boldsymbol{s}),
\]
such that for any novel position \(\boldsymbol{s}_{\text{new}} \notin \mathcal{P}\) (\(\mathcal{Q}\) for RX-moving), the prediction \(\mathcal{F}(\boldsymbol{s}_{\text{new}})\) closely approximates the true spatial spectrum.

\begin{figure}[t]
    \centering
   \includegraphics[width=0.99\linewidth]{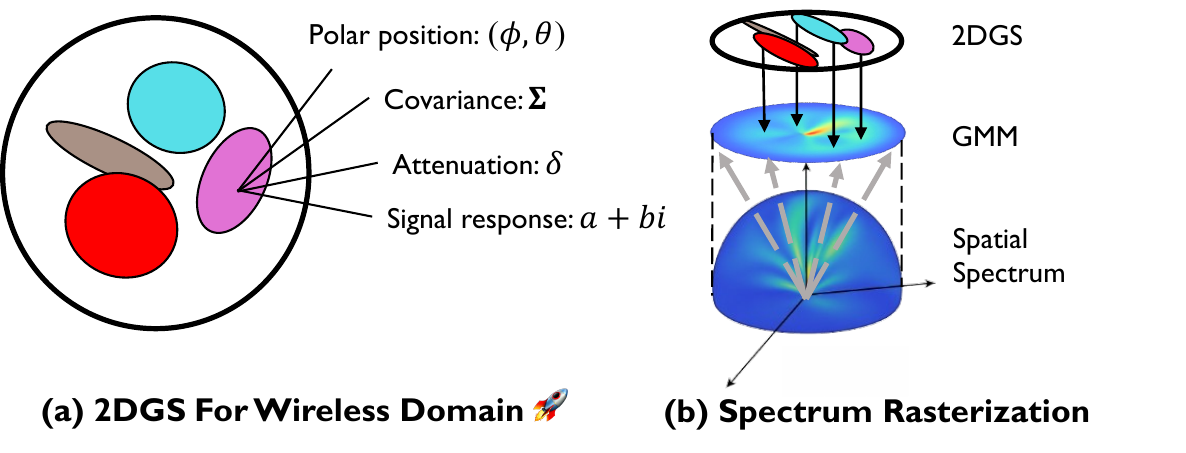}
    \caption{2DGS formulation and spectrum rasterization.}
    \label{2dgs}
\end{figure}

\section{Method}
In this section, we present the SwiftWRF architecture for modeling WRF under single-sided transceiver mobility. We first formulate a 2D Gaussian representation to construct the spatial spectrum, then introduce a deformation mechanism to capture their dynamics during transceiver mobility. Finally, we describe a two-stage training protocol and key optimization strategies. The following subsections detail each component.
\subsection{Modeling WRF Spectrum via 2DGS}

Under single-sided mobility, one transceiver is allowed to move, and a spatial spectrum of the WRF is collected at each position where it stops. To model a collected spectrum \(\mathbf{A}(\phi, \theta)\), we reformulate the 2D Gaussian formulation from Sec.~3.2 to the wireless domain and introduce a spatial rasterization process based on its original rasterization procedure.\\
\textbf{Wireless 2D Gaussian: }In the wireless domain, each 2D Gaussian primitive provides a localized representation of the spectrum signal, exhibiting a peak response at its center and decaying according to its associated Gaussian distribution. This distribution is parameterized by an angular center \(\boldsymbol{\mu}_n = (\phi, \theta)\) and a covariance matrix \(\boldsymbol{\Sigma}_n \in \mathbb{R}^{2 \times 2}\). To model the signal response, we introduce an attenuation factor \(\delta_n \in \mathbb{R}\) to control amplitude decay, and a signal vector \(\boldsymbol{\psi}_n \in \mathbb{R}^2\) to represent the real and imaginary parts of the complex-valued response. The angular center is normalized to lie within the valid spectrum range using a hyperbolic transformation
\[
\tilde{\theta} = \frac{\pi}{4}\left(\tanh(\theta) + 1\right), \quad \tilde{\phi} = \pi\left(\tanh(\phi) + 1\right).
\]
The covariance matrix \(\boldsymbol{\Sigma}_n\) is parameterized through its Cholesky factor \(\boldsymbol{l}_n\) (see Sec.~3.2). The attenuation factor is defined as \(\delta_n = \sigma(a_n)\), where \(a_n \in \mathbb{R}\) is a learnable scalar and \(\sigma(\cdot)\) denotes the sigmoid function. This gating mechanism regulates the contribution of each primitive to the reconstructed spectrum and helps prevent gradient explosion during optimization.\\
\textbf{Spectrum Rasterization: }
The localized signal representations described above are aggregated into a spectrum using a GMM framework. For each primitive, a Gaussian kernel weight \(\mathcal{K}\) is computed to modulate its signal response \(\boldsymbol{\psi}\), and the final spectrum value is obtained by summing the modulated responses from all primitives. Specifically, the kernel weight of the \(n\)-th primitive at the direction \((\phi, \theta)\) is computed as
\begin{equation}
\mathcal{K}(\boldsymbol{\mu}_n, \boldsymbol{\Sigma}_n, \delta_n) 
= \delta_n \cdot \exp\left( -\frac{1}{2} \boldsymbol{d}_n^\top \boldsymbol{\Sigma}_n^{-1} \boldsymbol{d}_n \right),
\end{equation}
where \(\boldsymbol{d}_n = (\phi - \phi_n,\, \theta - \theta_n)\) denotes the displacement between the direction \((\phi, \theta)\) and the primitive center \(\boldsymbol{\mu}_n\). The scalar \(\delta_n\) represents the attenuation factor of the primitive, while the exponential term \(\exp\left( -\frac{1}{2} \boldsymbol{d}_n^\top \boldsymbol{\Sigma}_n^{-1} \boldsymbol{d}_n \right)\) evaluates the likelihood that the sample at \((\phi, \theta)\) falls within the influence of that primitive. Utilizing the kernel to modulate each primitive's signal response \(\boldsymbol{\psi}_n\), the spectrum value at \((\phi, \theta)\) is computed as
\begin{equation}
\textbf{A}(\phi, \theta) = \sum_{n \in \mathcal{N}} \boldsymbol{\psi}_n \cdot \mathcal{K}(\boldsymbol{\mu}_n, \boldsymbol{\Sigma}_n, \delta_n).
\end{equation}
Drawing inspiration from differentiable rasterization, we refer to this process as \emph{spectrum rasterization}, as shown in Fig. \ref{2dgs}. The 2D Gaussian parameters are optimized iteratively through gradient backpropagation, guided by the loss between the predicted and ground-truth spectra. The entire pipeline is implemented in CUDA, with the spectrum domain partitioned into parallel tiles to fully exploit GPU acceleration.

\subsection{Deforming 2DGS for Mobility Modeling}

\begin{figure}[t]
    \centering
   \includegraphics[width=0.89\linewidth]{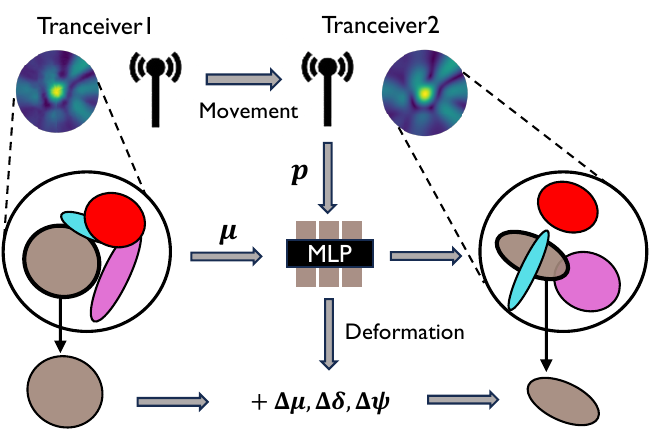}
    \caption{2DGS deformation for spectrum rendering under single-sided transceiver mobility.}
    \label{deformstraucture}
\end{figure}

The spectrum rasterization described in Sec.~4.1 defines a one-to-one mapping between the spatial spectrum domain and the 2DGS domain. As the transceiver moves across positions \(\boldsymbol{s} \in \mathcal{S}\), the resulting spectra  \(\{\mathbf{A}^{\boldsymbol{s}}\}_{\boldsymbol{s} \in \mathcal{S}}\) can be equivalently represented as a \textit{2DGS video} \(\{\mathcal{G}^{\boldsymbol{s}}\}_{\boldsymbol{s} \in \mathcal{S}}\), where each frame $\mathcal{G}^{\boldsymbol{s}}$ consists of a set of 2D Gaussians
\begin{equation}
\mathcal{G}^{\boldsymbol{s}} = \bigl\{(\boldsymbol\mu_n^{\boldsymbol{s}},\;\boldsymbol{\Sigma}_n,\;\boldsymbol\psi_n^{\boldsymbol{s}},\;\delta_n^{\boldsymbol{s}})\bigr\}_{n=1}^N,
\quad \boldsymbol{s} \in \mathcal{S}.
\end{equation}
We assume a constant number of primitives \(N\) across all frames and note that applying spatial rasterization to \(\mathcal{G}^{\boldsymbol{s}}\) yields the corresponding spectrum \(\mathbf{A}^{\boldsymbol{s}}\). To avoid the redundant per-frame 2DGS modeling, a Group-of-Pictures (GOP) paradigm is proposed: a global parameter set is first constructed as a \emph{reference frame} \(\mathcal{G}^0\), and each subsequent frame is represented by primitive-wise residuals
\begin{equation}
\Delta\mathcal{G}^{\boldsymbol{s}} = \bigl\{(\Delta\boldsymbol\mu_n^{\boldsymbol{s}},\;\Delta\boldsymbol\psi_n^{\boldsymbol{s}},\;\Delta\delta_n^{\boldsymbol{s}})\bigr\}_{n=1}^N.
\end{equation}
The primitive set at position \(\boldsymbol{s}\) is then given by
\begin{equation}
\mathcal{G}^{\boldsymbol{s}} = \bigl\{(\boldsymbol\mu_n^0 + \Delta\boldsymbol\mu_n^{\boldsymbol{s}},\;\boldsymbol{\Sigma}_n,\;\boldsymbol\psi_n^0 + \Delta\boldsymbol\psi_n^{\boldsymbol{s}},\;\delta_n^0 + \Delta\delta_n^{\boldsymbol{s}})\bigr\}_{n=1}^N,
\end{equation}
where \(\Delta\boldsymbol{\mu}_n^{\boldsymbol{s}}\), \(\Delta\boldsymbol{\psi}_n^{\boldsymbol{s}}\), and \(\Delta\delta_n^{\boldsymbol{s}}\) denote the residuals in position, response, and attenuation for the \(n\)th Gaussian primitive at moving location \(\boldsymbol{s}\). The residuals on the covariance matrices are omitted to preserve the triangular structure required by Cholesky factorization. Therefore, the spectrum rasterization under single-sided mobility generalizes to
\begin{equation}
\textbf{A}^{\boldsymbol{s}}(\phi, \theta) = 
\sum_{n \in \mathcal{N}} 
\left(\boldsymbol{\psi}_n^0 + \Delta \boldsymbol{\psi}_n^{\boldsymbol{s}}\right) \cdot 
\mathcal{K}\left(\boldsymbol{\mu}_n^0 + \Delta\boldsymbol{\mu}_n^{\boldsymbol{s}},\, \boldsymbol{\Sigma}_n^0,\, \delta_n^0 + \Delta \delta_n^{\boldsymbol{s}}\right).
\label{eq:polar-raster}
\end{equation}
However, modeling spectrum variations as residuals captures only discrete offsets at sampled positions \(\boldsymbol{s} \in \mathcal{S}\), and fails to generalize to unseen locations \(\boldsymbol{s}_{\text{new}} \notin \mathcal{S}\). To overcome this limitation, we introduce a continuous deformation field\footnote{We use “deformation” to denote continuous variation and “residual” for discrete offsets.}, implemented as a lightweight MLP that maps any moving position to the corresponding primitive residuals. Concretely, it predicts residuals for three components: First, \(\Delta\boldsymbol{\mu}_n^{\boldsymbol{s}}\) represents the motion vector of each primitive, capturing translational shifts caused by transceiver mobility. Second, \(\Delta\boldsymbol{\psi}_n^{\boldsymbol{s}}\) refines the signal response through a residual map that encodes fine-scale variations. Third, \(\Delta\delta_n^{\boldsymbol{s}}\) adjusts the attenuation factor to account for phenomena such as path blockage or reappearance. The deformation MLP takes the positional encodings of each primitive center \(\boldsymbol{\mu}\) and the moving position \(\boldsymbol{s}\) as input, and predicts their residuals as
\begin{equation}
\Delta\boldsymbol{\mu}_n^{\boldsymbol{s}}, \;\Delta \boldsymbol{\psi}_n^{\boldsymbol{s}}, \; \Delta \delta^{\boldsymbol{s}}_n 
= \mathcal{F}_\theta \big( \gamma(\boldsymbol{\mu}_n),\, \gamma(\boldsymbol{s}) \big),
\end{equation}
where \(\mathcal{F}_\theta\) is the deformation MLP with parameters \(\theta\), and \(\gamma(\cdot)\) denotes the positional encoding function. Given an input vector \(\boldsymbol{s} \in \mathbb{R}^d\), the embedding is defined as:
\begin{equation}
\gamma(\boldsymbol{s}) = \left[
\boldsymbol{s},\;
\{\sin(2^k \pi \boldsymbol{s})\}_{k=0}^{L-1},\;
\{\cos(2^k \pi \boldsymbol{s})\}_{k=0}^{L-1}
\right].
\label{eq:posenc}
\end{equation}
Here, each sine and cosine function is applied element-wise to all dimensions of \(\boldsymbol{s}\), and \(L\) denotes the number of frequency bands. This yields a total embedding dimension of \(d(2L + 1)\). Introducing multiple frequency bands allows the MLP to capture both coarse and fine-grained variations in the WRF.

\subsection{SwiftWRF}
As illustrated in Fig.~\ref{deformstraucture}, combining the 2D Gaussian formulation with the deformation MLP yields our complete model, SwiftWRF. In this unified framework, each primitive and the transceiver position are jointly input to the deformation MLP, which outputs per-primitive offsets. The deformed primitives are then passed through a CUDA-accelerated spatial rasterization pipeline to synthesize the target spectrum. This design integrates the efficiency of GS with the flexibility of a learnable continuous deformation field, enabling fast and accurate end-to-end WRF reconstruction without relying on volumetric grids.

\begin{figure}[t]
    \centering
    \includegraphics[width=0.99\linewidth]{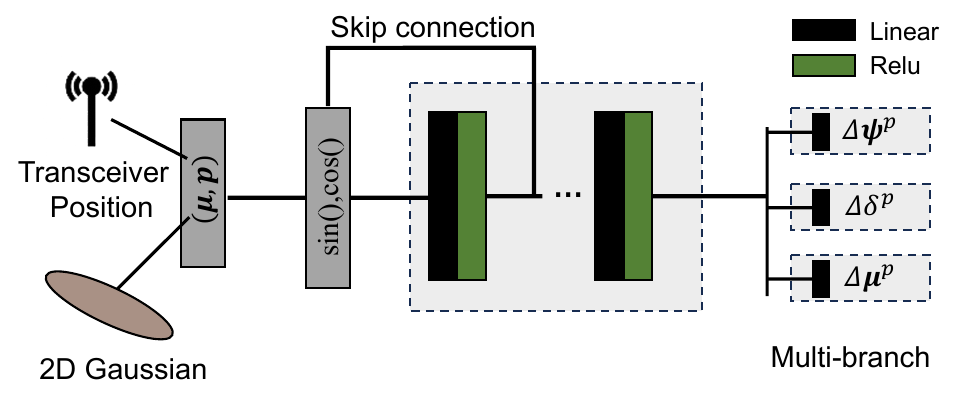}
    \caption{Structure of the deformation MLP.}
    \label{mlp}
\end{figure}

\begin{figure*}
    \centering
    \includegraphics[width=0.99\linewidth]{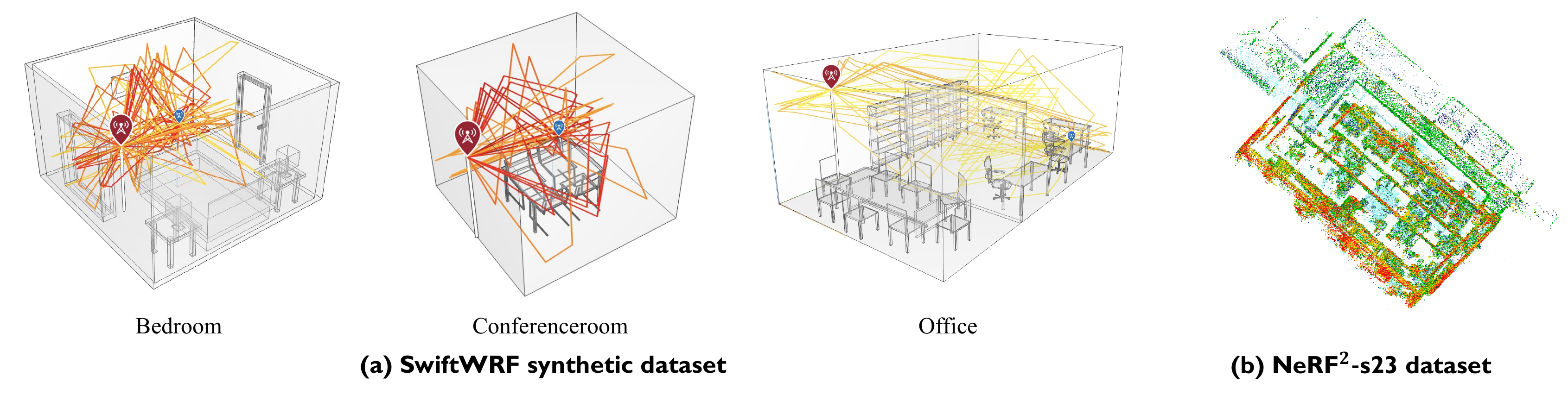}
    \caption{Geometry overview of the real-world (b) and synthetic dataset scenes (a).}
    \label{fig:dataset}
\end{figure*}

\subsection{Training Protocol}
\textbf{Coarse to fine training: }SwiftWRF adopts a two-stage coarse-to-fine training protocol, which is more efficient than training the full model from scratch. In the coarse stage, only the 2D Gaussian primitives are optimized to fit the entire set of measured spectra, while the deformation network remains frozen. This stage establishes a shared representation for the reference 2DGS frame and converges within seconds. In the fine stage, the deformation MLP is activated to learn adjustments conditioned on the location of the moving transceiver. To stabilize training, gradients to the primitive positions are frozen, concentrating optimization on the signal response variations.\\
\textbf{Loss function: } The goal of the training protocol is to reconstruct the spectrum at each moving position by minimizing the discrepancy between the predicted and ground-truth spectra. To achieve this, we use a hybrid loss function that combines the \(\ell_1\) loss with the Structural Similarity Index Measure (SSIM) loss:
\begin{equation}
\mathcal{L} = \lambda_1 \cdot \|\hat{\mathbf{A}}^{\boldsymbol{s}} - \mathbf{A}^{\boldsymbol{s}}\|_1 + (1 - \lambda_1) \cdot \big(1 - \text{SSIM}(\hat{\mathbf{A}}^{\boldsymbol{s}}, \mathbf{A}^{\boldsymbol{s}})\big),
\end{equation}
where \(\hat{\mathbf{A}}^{\boldsymbol{s}}\) is the reconstructed spectrum at position \(\boldsymbol{s}\), \(\mathbf{A}^{\boldsymbol{s}}\) is the ground-truth spectrum, and \(\lambda_1\) is a scalar weight balancing the contributions of the two terms. The \(\ell_1\) loss promotes pixel-wise accuracy, while the SSIM term encourages local structural similarity, ensuring both fidelity and spatial consistency in WRF reconstruction.\\
\textbf{Anneal Smoothing:} The training loss defined in the previous subsection does not impose any smoothness across different positions, which can easily lead to overfitting, i.e., poor results when reconstructing spectra at unseen locations. To address this, we adopt an anneal smoothing strategy~\cite{yang2024deformable}, where random Gaussian noise \(\boldsymbol{n}_i\) is added to the position $\boldsymbol{s}$ at each iteration \(i\) and gradually decays over the course of training. With slight abuse of notation, we denote the deformation components by \(\boldsymbol{\Delta}\) after applying anneal smoothing. The procedure is formulated as:
\begin{align}
\boldsymbol{\Delta} &= \mathcal{F}_\theta\big(\gamma(\boldsymbol{\mu}), \gamma(\boldsymbol{s} + \boldsymbol{n}_i)\big), \\
\boldsymbol{n}_i &= \gamma \cdot \mathcal{N}^3(0,1) \cdot \frac{2}{\sqrt[3]{N_s}} \cdot \left(1 - \frac{i}{\tau}\right).
\end{align}
Here, \(\gamma\) is a scaling factor, and \(\mathcal{N}^3(0,1)\) denotes a standard 3D normal distribution. The term \(\frac{2}{\sqrt[3]{N_s}}\) approximates the sampling interval, assuming that the \(N_s\) positions are uniformly distributed within a normalized unit cube. The parameter \(\tau\) denotes the iteration threshold for annealing. Introducing position smoothness improves SwiftWRF's generalization in the early stages of training, while gradually reducing its influence helps prevent excessive smoothing in later stages. Meanwhile, this regularization mitigates jitter artifacts during position interpolation by the deformation MLP.

\section{Synthetic spectra via ray-tracing}
We employ the ray tracing module from MATLAB’s Communications Toolbox \cite{matlab2023a,matlab2023b} to generate a corpus of spatial spectra across three representative indoor environments \cite{newrf}: \textit{conference room, office, and bedroom} (see Fig. \ref{fig:dataset} (a)), each simulated under two single-sided transceiver mobility configurations: 1) the TX is fixed and RX is placed at multiple positions to collect spectra, and 2) the RX is fixed to receive spectra from TX placed at multiple positions. These two configurations are among the most common in wireless communication and reflect typical use cases.

The ray-tracing module employs the shooting and bouncing rays (SBR) method to simulate realistic wireless propagation. For each scene, it launches rays from a TX toward its RX location in the scene's geometry model (e.g., mesh or CAD files) and computes specular reflections, diffraction, scattering, and path loss for each valid propagation path. The RX records channel responses, including AoA and multipath components, for the purpose of spectrum measurement. For each path, the phase shift is computed using Eq.\eqref{eq:phase shift}, and the overall channel is synthesized via superposition as described in Eq.\eqref{eq:received signal}. Finally, directional beam scanning is performed by sweeping across the angular directions defined in Eq.~\eqref{eq:beam} to construct the spatial spectrum. Depending on the configuration, we place either the TX or the RX through multiple positions and repeat the SBR process, producing about 6,000 spectra per scene.

\begin{table}[t]
  \centering
  \caption{Hyperparameter setting}
  \label{hyper}
  \begin{tabular}{lcc}
    \toprule
    Symbol & Meaning & Value \\
    \midrule
    $\gamma$      & Anneal smoothing weight  & 1 \\
    $\tau$ & Anneal smoothing threshold & 10k \\
    $\epsilon_g $  & Gaussian learning rate  &  1e-2  \\
    $\epsilon_m $  & MLP learning rate  & 8e-3 \\
    $\lambda_1$     & l1 loss weight & 0.7 \\
    \bottomrule
  \end{tabular}
\end{table}

\begin{table*}[ht]
\centering
\caption{Comparison of methods across different scenes, reported with median PSNR (↑), SSIM (↑), and mean L1 (↓). The best result is highlighted in dark gray, and the second best in light gray}
\label{per-scene}
\begin{tabular}{l|ccc|ccc|ccc|ccc}
\toprule
Method 
  & \multicolumn{3}{c|}{Conferenceroom (moving TX)} 
  & \multicolumn{3}{c|}{Bedroom (moving TX)} 
  & \multicolumn{3}{c|}{Office (moving TX)} 
  & \multicolumn{3}{c}{NeRF$^{2}$-s23} \\
\cmidrule(lr){2-13}  
& PSNR↑ & SSIM↑ & L1↓ 
& PSNR↑ & SSIM↑ & L1↓ 
& PSNR↑ & SSIM↑ & L1↓ 
& PSNR↑ & SSIM↑ & L1↓ \\
\midrule
RayTracing  & N/A & N/A & N/A   & N/A & N/A & N/A   & N/A & N/A & N/A   & - & 0.3300 & - \\
VAE   & 21.54 & 0.7095 & 0.0656  & 22.90 & 0.7468 & 0.0612  & 23.51 & 0.7647 & 0.0549  & 18.62 & 0.7133 & 0.1033 \\
DCGAN & 22.41 & 0.7290 & 0.0600  & 21.66 & 0.7314 & 0.0625  & 22.55 & 0.7441 & 0.0559  & 18.81 & 0.7189 & 0.0967 \\
$\text{NeRF}^2$ 
       & \cellcolor{secondgray}{26.12} & \cellcolor{secondgray}{0.8122} & 0.0555  
       & \cellcolor{secondgray}{25.99} & \cellcolor{secondgray}{0.8004} & 0.0608  
       & \cellcolor{secondgray}{27.52} & \cellcolor{secondgray}{0.8118} & \cellcolor{secondgray}{0.0497}  
       & \cellcolor{secondgray}{21.76} & \cellcolor{secondgray}{0.7732} & \cellcolor{secondgray}{0.0712}  \\
WRF-GS& 25.53 & 0.7888 & \cellcolor{secondgray}{0.0514}  
      & 26.10 & 0.7995 & \cellcolor{secondgray}{0.0534}  
      & 24.00 & 0.7422 & 0.0534  
      & 20.72 & 0.7588 & 0.0788 \\
Ours  & \cellcolor{bestgray}{30.75} & \cellcolor{bestgray}{0.9039} & \cellcolor{bestgray}{0.0410}  
      & \cellcolor{bestgray}{30.17} & \cellcolor{bestgray}{0.9166} & \cellcolor{bestgray}{0.0442}  
      & \cellcolor{bestgray}{32.96} & \cellcolor{bestgray}{0.9376} & \cellcolor{bestgray}{0.0355}  
      & \cellcolor{bestgray}{24.75} & \cellcolor{bestgray}{0.8714} & \cellcolor{bestgray}{0.0582} \\
\midrule
Method 
  & \multicolumn{3}{c|}{Conferenceroom (moving RX)} 
  & \multicolumn{3}{c|}{Bedroom (moving RX)} 
  & \multicolumn{3}{c|}{Office (moving RX)} 
  & \multicolumn{3}{c}{Average} \\
& PSNR↑ & SSIM↑ & L1↓ 
& PSNR↑ & SSIM↑ & L1↓ 
& PSNR↑ & SSIM↑ & L1↓ 
& PSNR↑ & SSIM↑ & L1↓ \\
\midrule
RayTracing  & N/A & N/A & N/A   & N/A & N/A & N/A   & N/A & N/A & N/A   & N/A & N/A & N/A \\
VAE     & 23.93 & 0.7664 & 0.0524  & 22.57 & 0.7414 & 0.0609  & 22.75 & 0.7562 & 0.0612 & 22.26 & 0.7426 & 0.0656 \\
DCGAN   & 21.26 & 0.7164 & 0.0608  & 21.79 & 0.7208 & 0.0601  & 20.95 & 0.7809 & 0.0529 & 21.35 & 0.7345 & 0.0641 \\
$\text{NeRF}^2$   
        & \cellcolor{secondgray}{28.21} & \cellcolor{secondgray}{0.8528} & \cellcolor{secondgray}{0.0387}  
        & \cellcolor{secondgray}{25.51} & \cellcolor{secondgray}{0.8057} & 0.0505  
        & 24.16 & \cellcolor{secondgray}{0.7863} & 0.0515
        & \cellcolor{secondgray}{25.61} & \cellcolor{secondgray}{0.8203} & \cellcolor{secondgray}{0.0539}  \\
WRF-GS  & 26.64 & 0.7916 & 0.0423  
        & 25.50 & 0.7901 & \cellcolor{secondgray}{0.0482}  
        & \cellcolor{secondgray}{24.07} & 0.7460 & \cellcolor{secondgray}{0.0510}
        & 24.65 & 0.7739 & 0.0541 \\
Ours    & \cellcolor{bestgray}{30.54} & \cellcolor{bestgray}{0.9137} & \cellcolor{bestgray}{0.0315}  
        & \cellcolor{bestgray}{28.15} & \cellcolor{bestgray}{0.9023} & \cellcolor{bestgray}{0.0398}  
        & \cellcolor{bestgray}{28.01} & \cellcolor{bestgray}{0.8755} & \cellcolor{bestgray}{0.0406}  
        & \cellcolor{bestgray}{29.33} & \cellcolor{bestgray}{0.9040} & \cellcolor{bestgray}{0.0417} \\
\bottomrule
\end{tabular}
\end{table*}

\section{Experiment}
\subsection{Implementation Details}
We evaluate SwiftWRF on both real‐world and synthetic datasets, and demonstrate its versatility on two case studies: AoA prediction and RSSI prediction. Spectrum rasterization is implemented on top of the open-source Gsplat framework \cite{ye2025gsplat} that allows CUDA-based parallel processing. The real and imaginary parts of each spatial spectrum value are treated as separate channels during rasterization and are decomposed into magnitude and phase. We initialize the canonical space with 10k Gaussian primitives, and randomly distribute them in 2D spectrum space. Training follows a two-stage schedule: a coarse stage of 10k epochs to learn the canonical 2DGS representation, followed by a fine-tuning stage of 100k–150k epochs to optimize the deformation MLP. The deformation network is an eight-layer MLP with skip connections every two layers and a hidden width of 156 (See Fig. \ref{mlp}); positional encodings use 10 frequency bands for the primitive center and 6 for the moving transceiver position. All components are implemented in PyTorch and trained on an NVIDIA RTX 3090 GPU. Detailed hyperparameters are reported in Table \ref{hyper}.

\subsection{Dataset}
We evaluate our proposed framework on two types of datasets. The first is an open-source real-world dataset~\cite{Nerf2} collected in a laboratory environment equipped with tables, shelves, and partitioned rooms. A \(4\times4\) antenna array operating at 915 MHz functions as the RX, while a passive RFID tag continuously transmits RN16 messages as the TX. The RX remains stationary, and the TX is systematically moved throughout the environment to capture varied spatial spectra. The second is our synthetic spatial spectrum dataset, which consists of three representative indoor scenes where the RX is also equipped with a \(4\times4\) antenna array. Each scene contains two configurations under single-sided mobility. The real-world dataset provides a large-scale evaluation scenario with more complex geometric structures, whereas the synthetic datasets feature simpler environments, as illustrated in Fig.~\ref{fig:dataset}.

\begin{figure*}[t]
    \centering
    \includegraphics[width=0.99\linewidth]{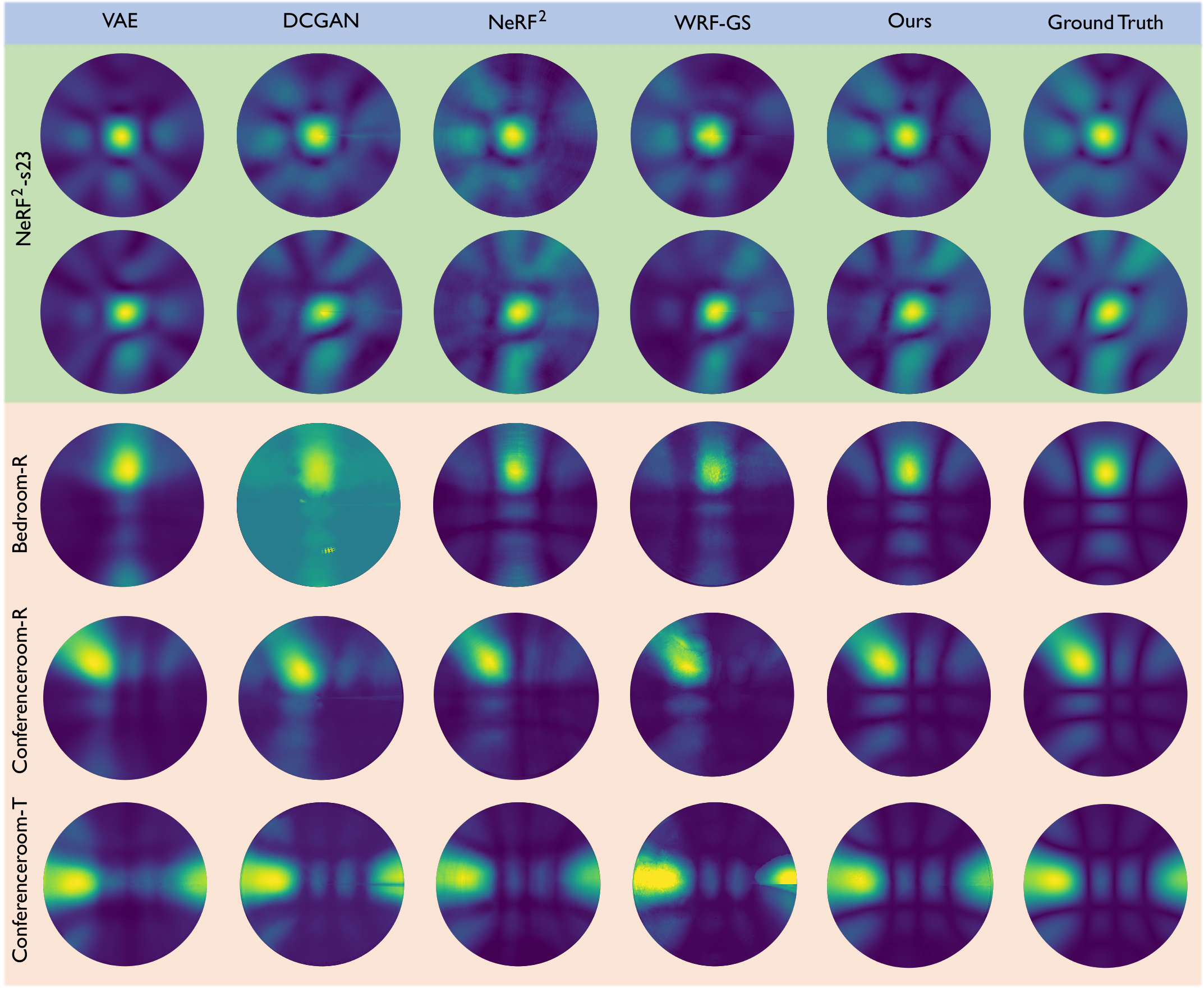}
    \caption{Spectrum visualizations on selected scenes, including two positions from the NeRF²-23 public dataset and one position each from bedroom (RX-side moving), conference room (RX-side moving), and conference room (TX-side moving) in our dataset. Rendered at \(90 \times 360\)); please zoom in to see details.}
    \label{spectrumvis}
\end{figure*}

\begin{figure}[t]
    \centering
    \includegraphics[width=0.91\linewidth]{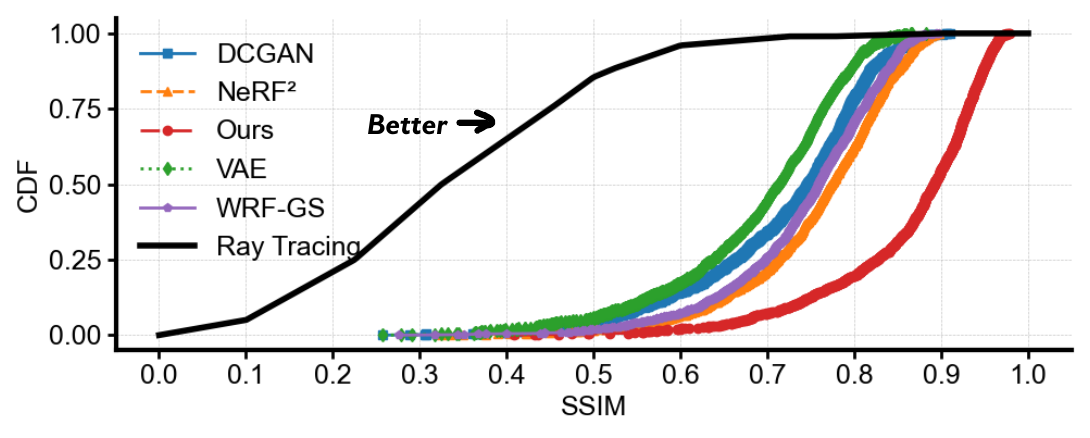}
    \vspace{-0.4cm}
    \caption{CDF of SSIM on $\text{NeRF}^2$-s23 dataset.}
    \vspace{-0.4cm}
    \label{cdfssim}
\end{figure}

\subsection{Benchmark}
We evaluate SwiftWRF against a representative set of benchmarks, including the simulation-based ray-tracing method, deep-learning approaches (VAE and DCGAN), the implicit-neural-representation model $\text{NeRF}^2$, and the 3DGS solution WRF-GS.
\begin{itemize}
    \item \textbf{Ray Tracing \cite{raytracing}:} Ray tracing emits rays from the TX through each pixel of the scene model reconstructed from the point cloud to recursively model wireless propagation. We use MATLAB’s RayTracing toolkit to perform this simulation with high physical fidelity.
    \item \textbf{Variational Autoencoder (VAE) \cite{vae}}: A VAE learns to map observed data into a continuous latent space, from which new samples can be generated. In the context of WRF, the VAE is trained to predict the spatial spectrum at arbitrary TX locations based on discrete observations.
    \item \textbf{Deep Convolutional Generative Adversarial Network (DCGAN) \cite{gan}}: DCGAN consists of a convolutional generator and a convolutional discriminator trained in an adversarial loop. The generator aims to produce spectrum maps that resemble real measurements, while the discriminator learns to distinguish between real and synthesized outputs.
    \item \textbf{$\text{NeRF}^2$ \cite{Nerf2}}: $\text{NeRF}^2$ models the WRF as a learnable volumetric density field. It encodes the scene density and signal strength into an MLP to render spatial spectra at arbitrary TX locations.
    \item \textbf{WRF-GS: \cite{wrf}} WRF-GS is a GS-based method that leverages neural 3D Gaussian primitives for spectrum synthesis. It reconstructs spatial spectra at arbitrary TX positions by rasterizing learned Gaussians conditioned on TX location.
\end{itemize}

\subsection{Metrics}
We assess reconstruction quality with two complementary metrics. Peak signal-to-noise ratio (PSNR) \cite{korhonen2012peak} quantifies pixel-wise fidelity by measuring the logarithmic ratio between the maximum possible signal power and the mean squared error; it is widely used in signal-reconstruction tasks to gauge absolute error levels. The structural similarity index measure (SSIM) evaluates perceived image quality by comparing local patterns of luminance, contrast, and structure \cite{hore2010image}. We also report the L1 loss, which directly quantifies reconstruction accuracy by measuring the mean absolute per-pixel error. For computational efficiency, we report training time in hours and inference speed in seconds. Unless otherwise noted, all model parameters are stored in 32-bit floating point (FP32).

\begin{figure}[t]
    \centering
    \includegraphics[width=0.9\linewidth]{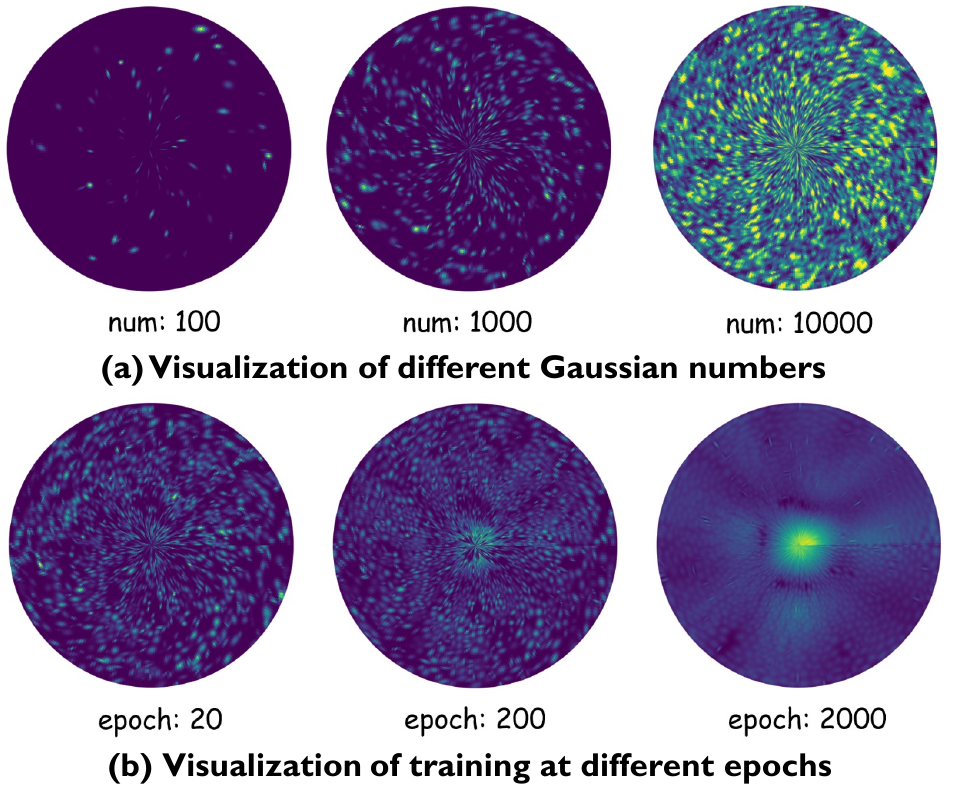}
    \caption{2D Gaussian visualization.}
    \label{fig:process}
\end{figure}

\subsection{WRF reconstruction}
To demonstrate the effectiveness of SwiftWRF, we first present per-scene evaluations against baseline methods across seven representative scenes. We then provide visual comparisons on selected examples to highlight qualitative differences. Finally, we report fast inference results on the $\text{NeRF}^2$ dataset and analyze the impact of other contributing factors.

\begin{table}[t]
\centering
\caption{Comparisons with recent GS-based methods on the NeRF$^2$-s23 dataset}
\label{gscompare}
\renewcommand{\arraystretch}{1.15}
\resizebox{\columnwidth}{!}{%
\begin{tabular}{lccc}
\toprule
\textbf{Algorithm} & \textbf{Training Time (hrs)}$\downarrow$ & \textbf{Rendering Time (ms)}$\downarrow$ & \textbf{Median SSIM}$\uparrow$ \\
\midrule
RFSPM   & N/A & 4.2 & N/A \\
GSpaRC  & \textbf{0.33} & 0.39 & 0.82 \\
\midrule
Ours    & 0.83 & \textbf{0.01} & \textbf{0.87} \\
\bottomrule
\end{tabular}%
}
\end{table}

\begin{table}[t]
\centering
\small
\caption{Comparison of model storage and training time to convergence}
\resizebox{0.9\linewidth}{!}{
\begin{tabular}{lcc}
\toprule
Method & Storage (MB) $\downarrow$ & Training time (to convergence) $\downarrow$ \\
\midrule
NeRF2   & 8.00  & 5.8 h \\
WRF-GS  & 18.15 & 1.5 h \\
Ours    & \textbf{3.47}  & \textbf{50 min} \\
\bottomrule
\end{tabular}}
\label{tab:storage_time}
\end{table}

1) \textit{Per-scene Evaluation:} Table~\ref{per-scene} summarizes the evaluation metrics comparing our method against all baselines across all evaluated scenes, which include the large-scale real-world dataset \(\text{NeRF}^2-\text{s}23\) and the customized synthetic scenes. Overall, SwiftWRF achieves a PSNR improvement of 2.33–10.41 dB over competing methods, with consistent gains also observed in SSIM and l1 loss. On the large-scale $\text{NeRF}^2-\text{s}23$ dataset, SwiftWRF achieves a PSNR improvement of 2.99–4.03 dB over both $\text{NeRF}^2$ and WRF-GS, and especially outperforms the ray tracing method with a 63\% increase in SSIM. This is attributed to the fact that ray tracing relies heavily on geometric priors, and the LiDAR-based scene reconstruction may be insufficiently accurate. Fig.~\ref{cdfssim} presents cumulative SSIM curves across all evaluated positions, demonstrating that our method outperforms all baselines by a substantial margin. On synthetic scenes with simpler layouts, SwiftWRF consistently achieves superior results, with a PSNR improvement ranging from 2.65~dB to 10.41~dB over all baselines. Notably, the comparable performance observed under both mobility scenarios suggests that SwiftWRF captures two distinct types of field variation: in the RX-moving case, the MLP likely models spatial variations across different receiving positions, whereas in the TX-moving case, it captures temporal variations at a fixed location.We also compare our method with recent GS-based approaches \cite{yang2025scalable, nukapotula2025gsparc} on the public NeRF$^2$-s23 dataset, as summarized in Table~\ref{gscompare}. Overall, our method achieves substantially faster rendering while maintaining comparable or better reconstruction quality. Notably, we reach a median SSIM of 0.85 within 0.33 hours, already surpassing GSpaRC’s reported converged performance. Since RFSPM does not release its code, we report comparisons only against the metrics provided in their paper.

\begin{figure}[t]
  \centering
  \begin{minipage}[t]{0.500\linewidth}
    \centering
    \includegraphics[width=\linewidth]{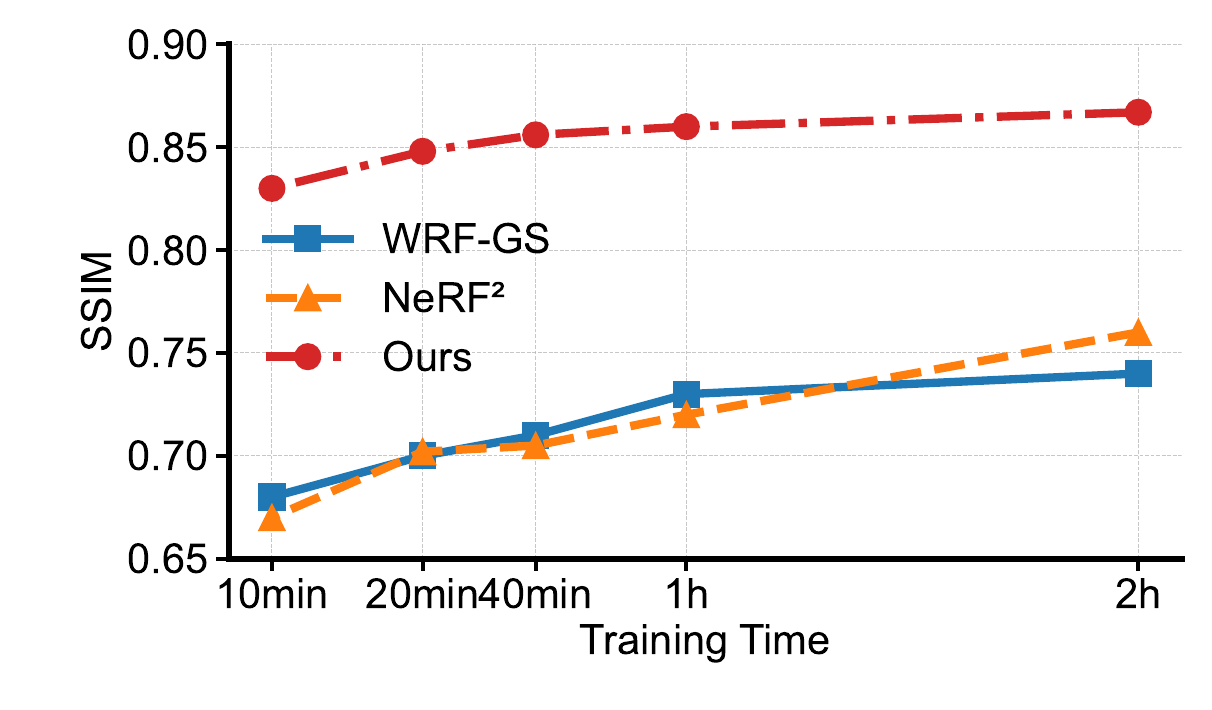}
    \small (a) Training time.
  \end{minipage}
  \hfill
  \begin{minipage}[t]{0.482\linewidth}
    \centering
    \includegraphics[width=\linewidth]{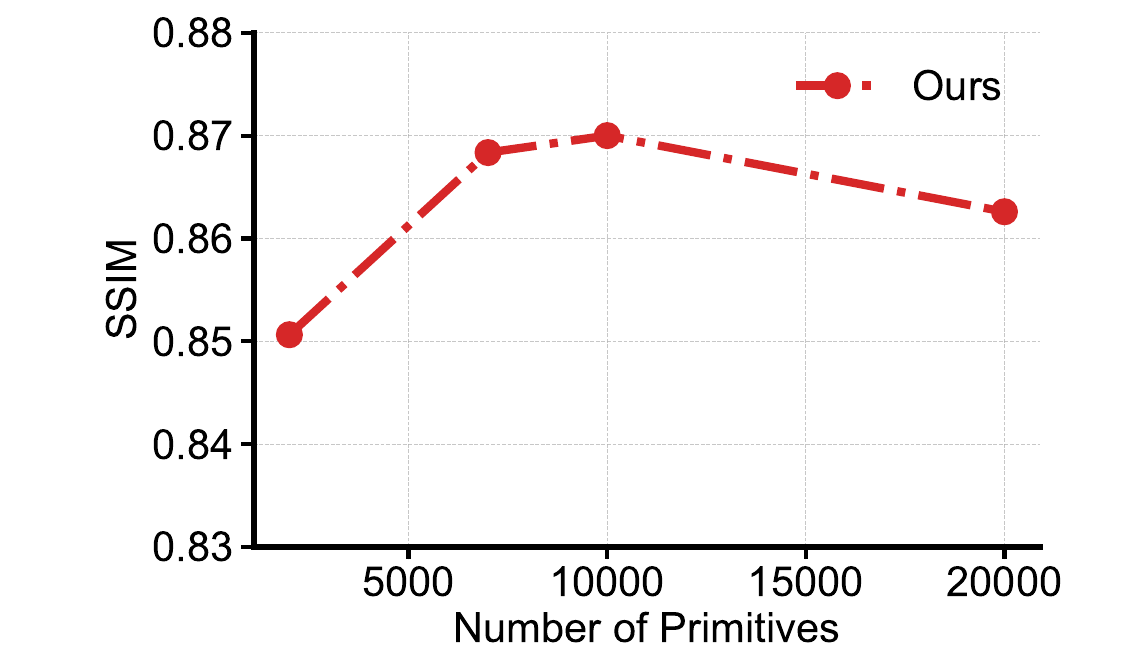}
    \small (b) Primitive count.
  \end{minipage}
  \caption{Median SSIM under different conditions: (a) training time, and (b) primitive count.}
  \label{factor}
\end{figure}
\begin{figure}[t]
    \centering
    \includegraphics[width=1\linewidth]{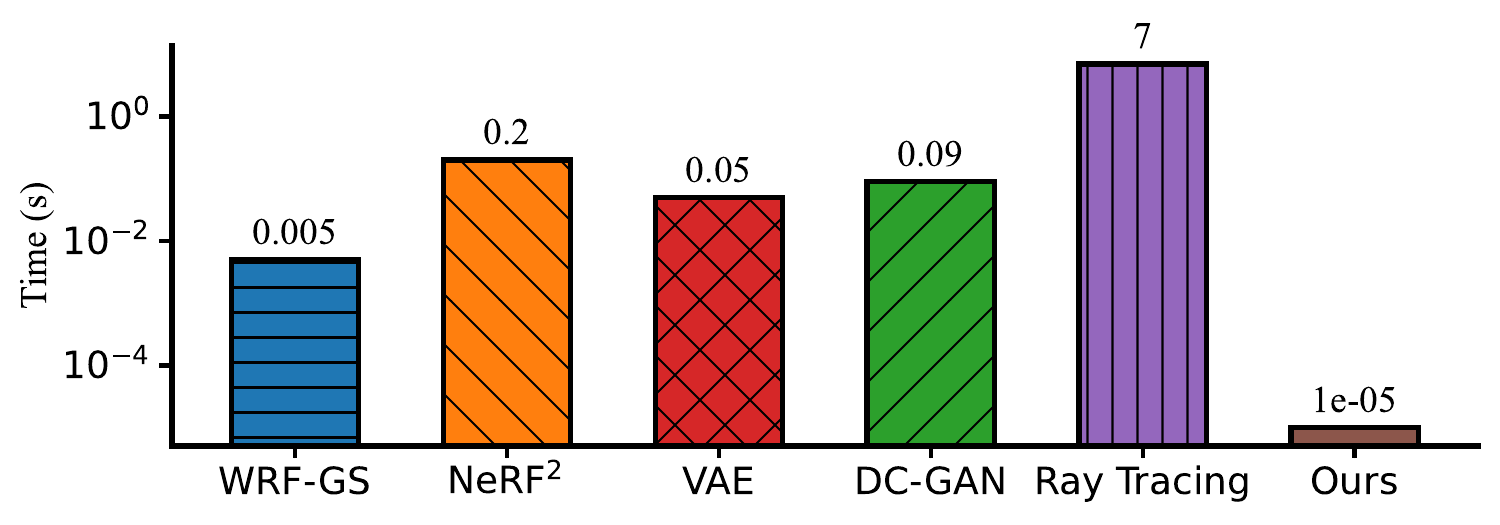}
    \vspace{-0.5cm}
    \caption{Comparison of inference time for a single spectrum.}
    \label{decode}
\end{figure}

\begin{figure}[t]
    \centering
    \includegraphics[width=0.94\linewidth]{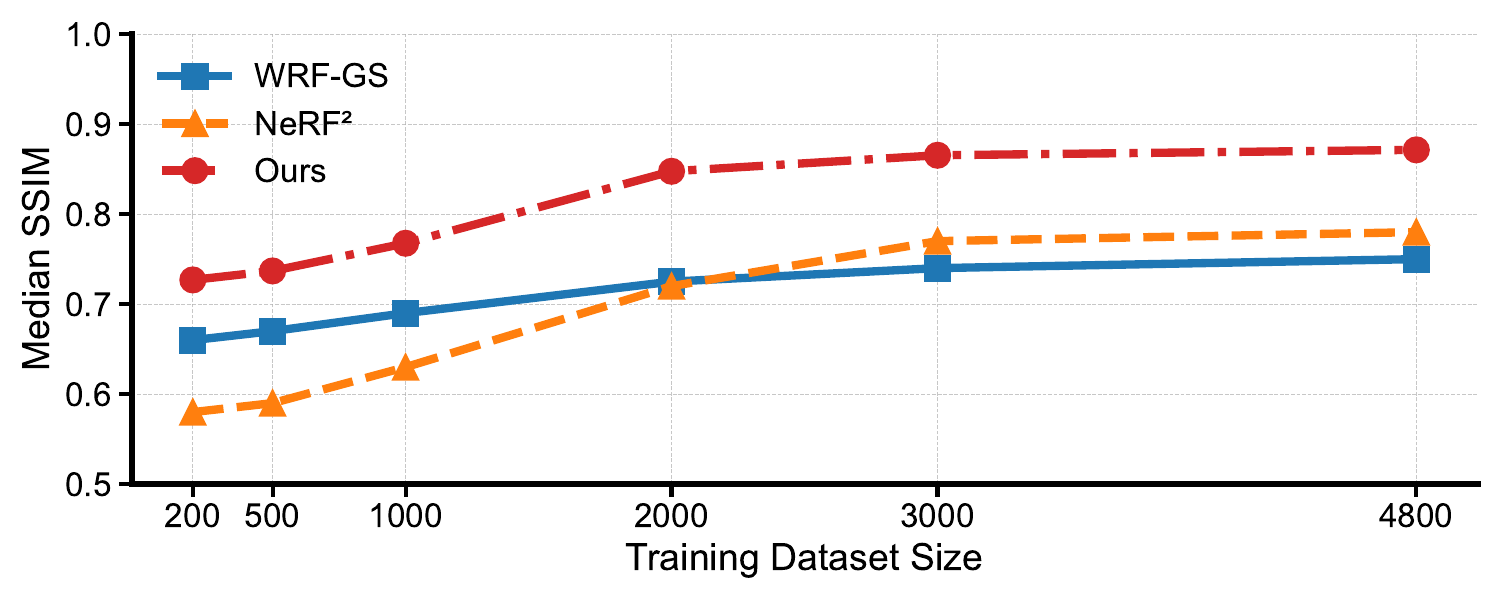}
    \caption{Evaluation on different training scales.}
    \label{Scale}
\end{figure}

\begin{figure*}[t]
    \centering
    \includegraphics[width=\linewidth]{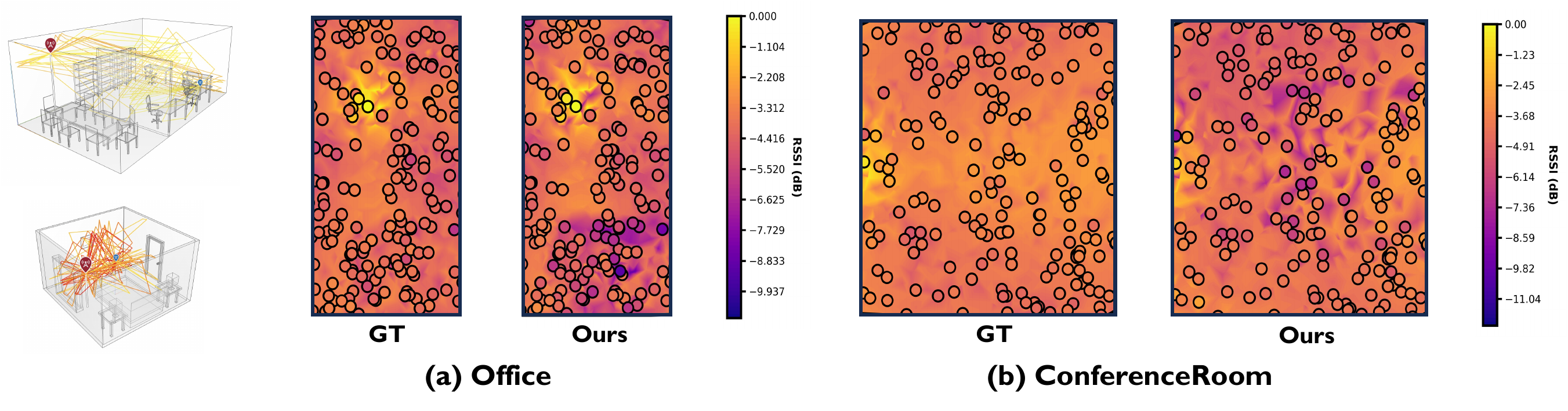}
    \caption{Wireless radiance field distribution visualization on \textit{Office} and \textit{ConferenceRoom}. Circles indicate a subset of training samples.}
    \label{fig:heatmap}
\end{figure*}

\begin{figure*}[t]
    \centering
    \includegraphics[width=\linewidth]{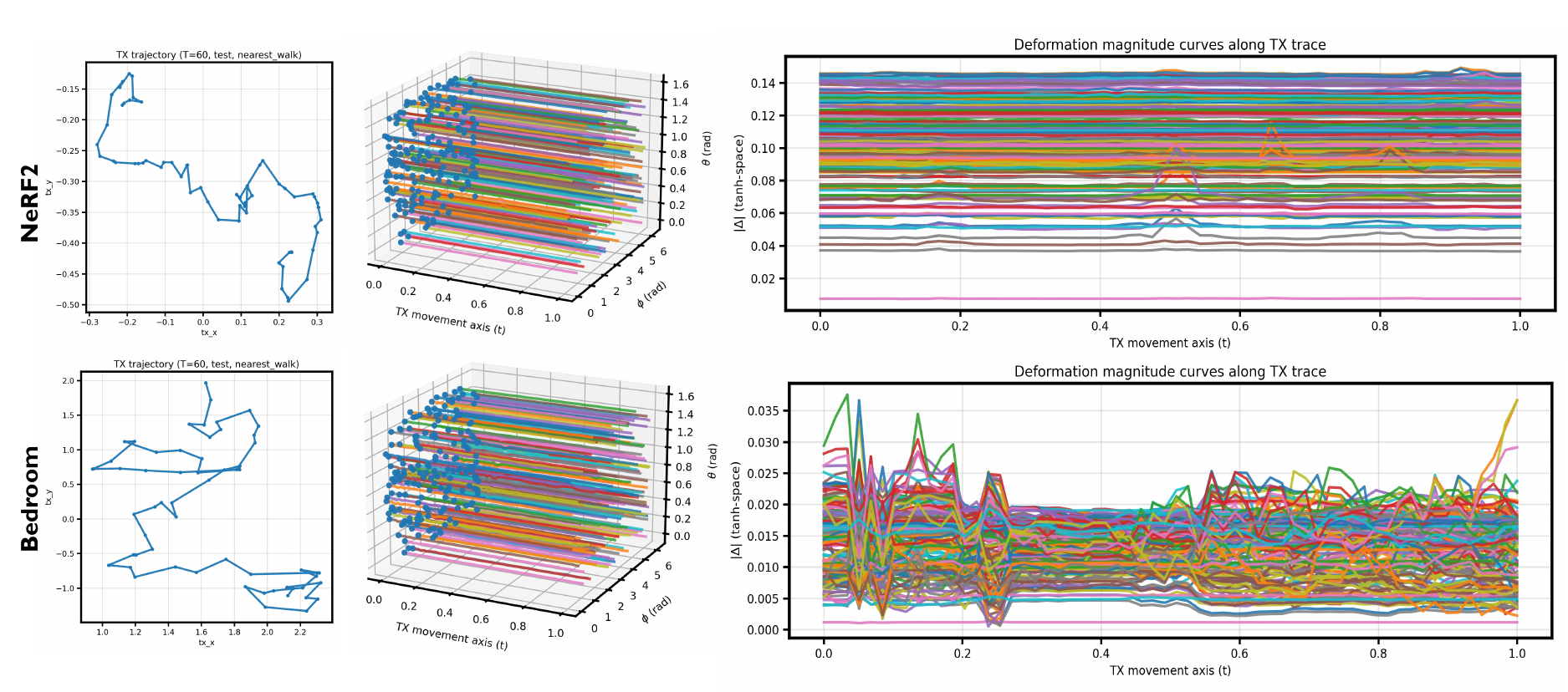}
    \caption{Toy example of 2DGS deformation. Left: a sample TX trajectory. Middle: per-Gaussian center deformation versus trajectory length. Right: deformation magnitude versus trajectory length.}
    \label{fig:defor_vis}
\end{figure*}

2) \textit{Visual comparison:} A few toy examples from the visual comparisons are presented in Fig.~\ref{spectrumvis} to highlight representative qualitative differences across methods. For clarity, we recommend zooming in to better observe reconstruction details. As shown, $\text{NeRF}^2$ produces ring-like blurring artifacts that degrade the overall signal spectrum. In contrast, WRF-GS exhibits localized blob-like noise patterns, likely caused by floaters introduced during the training of the 3DGS process. Meanwhile, traditional deep learning-based methods such as VAE and DCGAN yield heavily blurred results with limited structural fidelity. These visual observations further validate the superior quality of the signal maps generated by our method, in line with the objective metrics reported in Table~\ref{per-scene}.

3) \textit{Efficiency:} We evaluate both the training and inference efficiency of a single-spectrum reconstruction across all methods on the $\text{NeRF}^2-\text{s23}$ dataset, with results summarized in Fig.~\ref{decode} and Fig. \ref{factor} (a). All methods are executed on a single NVIDIA RTX 3090 GPU, except for ray tracing. As shown in Fig.~\ref{decode}, our method achieves a substantial performance advantage, reaching approximately 100k FPS—over 500× faster than WRF-GS and more than 20,000× faster than $\text{NeRF}^2$. In addition, our model consistently outperforms both WRF-GS and $\text{NeRF}^2$ during training, achieving around 20\% higher median SSIM across varying rendering times (see Fig. \ref{factor} (a)). This efficiency arises from our compact 2D Gaussian formulation, which enables highly parallelized CUDA-based rasterization without incurring the overhead of volumetric rendering in $\text{NeRF}^2$ or the Jacobian-based projection computations required by WRF-GS. Beyond its fast speed, SwiftWRF also delivers superior compactness, requiring only 1202 MB of GPU memory compared with 1804 MB for WRF-GS and 8978 MB for $\text{NeRF}^2$.

4) \textit{Scale with training data:} To assess adaptability to limited training data, we compare SwiftWRF with $\text{NeRF}^2$ and WRF-GS across varying training set sizes. Specifically, we partition the original training set into subsets of different sizes \([200, 500, 1000, 2000, 3000]\), and evaluate all models on the full test set to assess their robustness under limited data conditions. As shown in Fig.~\ref{Scale}, the performance of all methods begins to saturate once the training size exceeds 3000 samples. Across all scales, SwiftWRF consistently outperforms both $\text{NeRF}^2$ and WRF-GS in terms of SSIM, demonstrating superior generalization under low-data regimes.

5) \textit{Primitive count: }Another advantage of our framework is the ability to customize the number of primitives, allowing for flexible model scaling from lightweight to high-fidelity configurations, as illustrated in Fig.~\ref{factor}(b). Interestingly, increasing the primitive count beyond a certain threshold does not yield further quality improvements and may even lead to overfitting. 

6) \textit{Storage: }According to Table~\ref{tab:storage_time}, our method is the most compact among NeRF$^2$ and WRF-GS, reducing the model storage from 8.00\,MB (NeRF$^2$) and 18.15\,MB (WRF-GS) to 3.47\,MB, which corresponds to 56.6\% and 80.9\% less storage, respectively.

7) \textit{Resolution: }We set the spectrum resolution to $90 \times 360$ following NeRF$^2$’s evaluation protocol, but our method is not limited to this setting. Thanks to the explicit primitive-based formulation of Gaussian splatting, it can scale to higher resolutions by increasing the number of Gaussians. We additionally evaluate on \textit{Bedroom} at $0.5^\circ$ resolution ($180 \times 720$), and observe effective reconstruction at this higher resolution. On the test set, we report an L1 loss of $0.039$ and a median SSIM of $0.898$. Training is extended slightly from 50 minutes to 70 minutes.

\begin{table}[t]
\centering
\caption{Ablation study of deformation components}
\label{abla1}
\resizebox{\columnwidth}{!}{
\large
\begin{tabular}{
  >{\centering\arraybackslash}p{1.4cm}
  >{\centering\arraybackslash}p{2.2cm}
  >{\centering\arraybackslash}p{2.2cm}
  >{\centering\arraybackslash}p{2.2cm}
  >{\centering\arraybackslash}p{3.5cm}
}
\toprule
\textbf{$\Delta \boldsymbol{\mu}$}
  & \textbf{$\Delta \boldsymbol{\psi}$}
  & \textbf{$\Delta \delta$}
  & \textbf{$\Delta \boldsymbol{\Sigma}$}
  & \textbf{SSIM} $\uparrow$/\textbf{L1l} $\downarrow$ \\
\midrule
$\checkmark$ & $\checkmark$ & $\checkmark$ & \ding{55}      & 0.9040 / 0.0417\\
$\checkmark$ & $\checkmark$ & $\checkmark$ & $\checkmark$   & 0.7313 / 0.0777\\
$\checkmark$ & $\checkmark$ & \ding{55}    & \ding{55}      & 0.7618 / 0.0707\\
$\checkmark$ & \ding{55}    & $\checkmark$ & \ding{55}      & 0.8930 / 0.0424\\
\ding{55}    & $\checkmark$ & $\checkmark$ & \ding{55}      & 0.8735 / 0.0474\\
\bottomrule
\end{tabular}
}
\end{table}

\begin{table}[t]
  \centering
  \caption{Ablation study of training items}
  \label{abla2}
  \resizebox{0.99\columnwidth}{!}{%
    \large
    \begin{tabular}{
      >{\centering\arraybackslash}p{5cm}
      >{\centering\arraybackslash}p{4cm}
      >{\centering\arraybackslash}p{3cm}
    }
      \toprule
      \textbf{Ablation items}
        & \textbf{SSIM}$\uparrow$\,/\,\textbf{L1l}$\downarrow$
        & \textbf{Training Time (min)}$\downarrow$ \\
      \midrule
      N/A
        & 0.9040 / 0.0417
        & 50.0 \\
      w/o Stop Gradient
        & 0.9048 / 0.0497
        & 77.0 \\
      w/o Coarse Training
        & 0.9045 / 0.0366
        & 50.0 \\
      w/o Anneal Smoothing
        & 0.9025 / 0.0428
        & 48.0 \\
      w/o SSIM Loss
        & 0.8269 / 0.0529
        & 47.0 \\
      w/ L2 Loss
        & 0.8329 / 0.0518
        & 56.0 \\
      \bottomrule
    \end{tabular}%
  }
\end{table}

\subsection{Visualization and analysis}
To better illustrate the learned field and deformation behavior, we include two additional visualizations in the evaluation section. First, Fig.~\ref{fig:heatmap} reports the full-field distribution for \textit{Bedroom} and \textit{ConferenceRoom}. At each sampled position, we compute the RSSI by averaging the spectrum magnitude over both azimuth and elevation directions, and visualize the resulting ground-truth field together with the field reconstructed by our method. Second, Fig.~\ref{fig:defor_vis} visualizes the deformation of 2D Gaussian primitives along a continuous TX trajectory. We extract a sample trajectory from the two scenes using a nearest-walk strategy and plot the per-Gaussian center deformation as well as the corresponding deformation magnitude across 60 trajectory steps. Notably, the deformation patterns vary locally along the trajectory, indicating that our position-conditioned deformation field models the residuals at different locations in a largely independent manner.

\subsection{Ablation Study}
In this section, we conduct a comprehensive ablation study of the SwiftWRF design. To validate the effectiveness of each design choice, the ablation is organized into two parts: (1) deformation components and (2) training protocols.
For (1), as shown in Table~\ref{abla1}, applying deformation to all components does not yield the best performance as initially expected. In particular, adding covariance deformation leads to an 86.44\% increase in l1 loss, primarily due to the numerical instability it introduces by violating the lower triangular structure required for Cholesky decomposition. Excluding the covariance branch, all other deformation components contribute positively to performance, as verified through our step-wise ablation. Notably, position deformation proves to be the most critical, with its removal resulting in a 13.56\% increase in l1 loss. Signal deformation has a relatively minor impact on reconstruction accuracy but still provides measurable gains in overall performance. These findings highlight the complementary roles of each component and confirm the optimality of our deformation design. For (2), the results reported in Table~\ref{abla2} show that applying stop-gradient significantly reduces training time, with only a marginal drop in reconstruction performance. All other training configurations have relatively minor impact on training speed, except for the use of l2 loss, which introduces a 12.13\% overhead. The results also highlight that SSIM regularization and anneal smoothing reduce l1 loss to 2.72\% and 26.91\%, respectively, demonstrating their effectiveness in mitigating overfitting.

\begin{figure*}[t]
    \centering
    \includegraphics[width=0.99\linewidth]{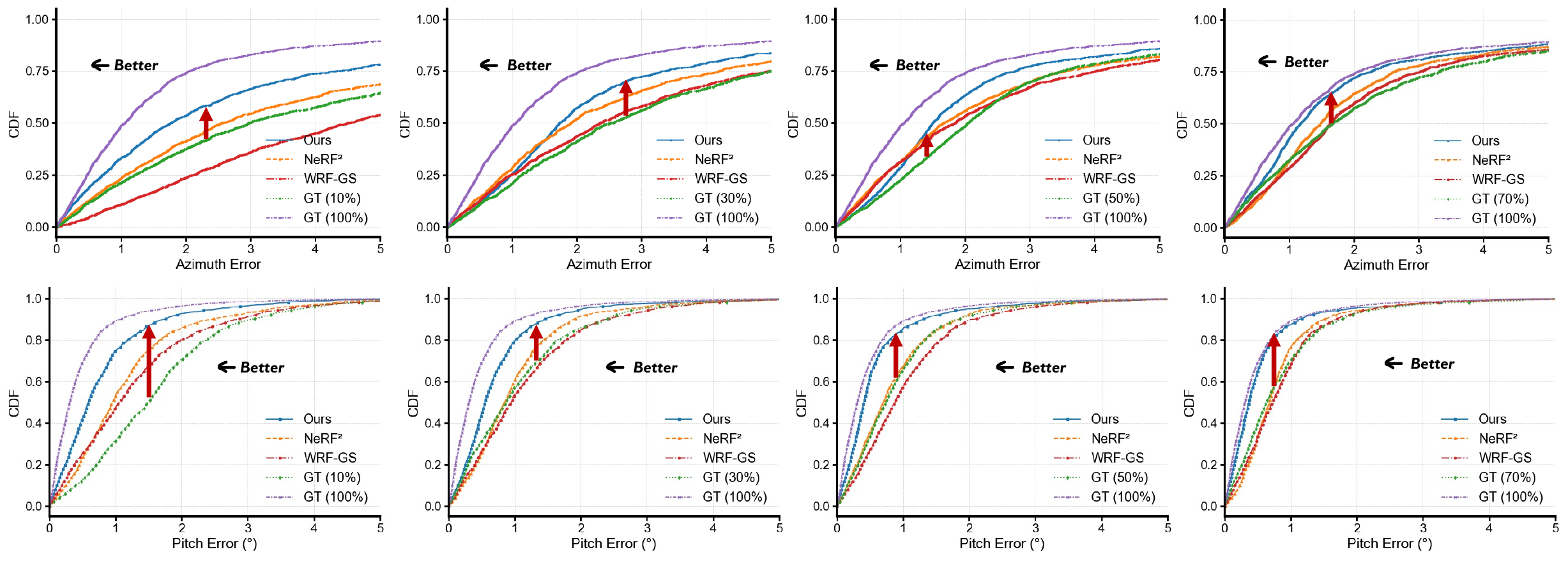}
    \caption{CDF of AoA prediction error under four different mixture ratios. The red arrow represents the performance improvement achieved by incorporating novel spectra synthesized by SwiftWRF.}
    \label{fig:aoa_cdf}
\end{figure*}

\section{Case Study}
In this section, we present two representative case studies in the context of channel parameter prediction to demonstrate SwiftWRF’s adaptability to multi-tasking. The first case is AoA prediction, where we adopt a turbo-learning manner to showcase SwiftWRF’s capability for data augmentation. The second task is RSSI prediction, for which we adapt the output head of SwiftWRF to directly regress RSSI values.
\subsection{AoA Prediction}
\textit{Case setup: }We perform AoA prediction using the Angular Artificial Neural Network (AANN) \cite{iArk}, which is designed to learn the mapping between spatial spectra and their corresponding AoA values. To demonstrate SwiftWRF’s potential in reducing the burden of data collection, we compare two training strategies: turbo learning and naïve learning. In the turbo learning setup, a portion of the ground-truth spectra is replaced with spectra generated by SwiftWRF, whereas in the naïve learning setup, the model is trained solely on the full set of ground-truth spectra. The AANN model is deployed using the ResNet-50 network \cite{he2016deep}, which is followed by an MLP for regression. The evaluation is conducted on the NeRF$^2$-s23 dataset. \\
\textit{Results: }Fig.~\ref{fig:aoa_cdf} presents the CDF of azimuth and pitch angle prediction error under four turbo learning mixture ratios: 10\%, 30\%, 50\%, and 70\%\footnote{The mixture ratio refers to the proportion of ground-truth spectra in the training set. For example, a 10\% ratio indicates that only 10\% of the samples are ground truth, while the remaining 90\% are generated spectra.}. The purple curve labeled as GT (100\%) represents the upper bound of prediction fidelity, obtained through the naïve training strategy described above. In contrast, the GT curves corresponding to each mixture ratio serve as theoretical lower bounds, as they retain only the specified proportion of the ground-truth training set. Fig.~\ref{fig:aoa_cdf} uses red arrows to highlight the performance gains from SwiftWRF-generated spectra, with improvements of 18\%, 21\%, 16\%, and 17\% across the four ratios. As reported in Table~\ref{tab:turbo_aoa}, we further assess SwiftWRF’s effectiveness by comparing its turbo learning performance against NeRF$^2$ and WRF-GS under the same mixture ratios. Across all mixture ratios, SwiftWRF achieves the lowest median prediction errors in both pitch and azimuth. In particular, it outperforms NeRF$^2$ and WRF-GS in azimuth prediction with relative improvements ranging from 8.7\%–30.6\% and 19.2\%–60.7\%, respectively.

\begin{table*}[t]
\centering
\caption{Median pitch and azimuth prediction errors (in degrees)}
\label{tab:turbo_aoa}
\setlength{\tabcolsep}{8pt} 
\begin{tabular}{c|cc|cc|cc}
\toprule
\textbf{Ratio} & \multicolumn{2}{c|}{\textbf{SwiftWRF}} & \multicolumn{2}{c|}{\textbf{NeRF$^2$}} & \multicolumn{2}{c}{\textbf{WRF-GS}} \\
& pitch angle ↓ & azimuth angle ↓ & pitch angle ↓ & azimuth angle ↓ & pitch angle ↓ & azimuth angle ↓ \\
\midrule
10\% & \underline{\textbf{0.5916}} & \underline{\textbf{1.7691}} & 0.9574 & 2.5479 & 1.0557 & 4.5064 \\
30\% & \underline{\textbf{0.5527}} & \underline{\textbf{1.7369}} & 0.8496 & 1.9031 & 0.9477 & 2.3639 \\
50\% & \underline{\textbf{0.4101}} & \underline{\textbf{1.4978}} & 0.7037 & 1.6628 & 0.8783 & 1.8541 \\
70\% & \underline{\textbf{0.3697}} & \underline{\textbf{1.1731}} & 0.6738 & 1.4782 & 0.7272 & 1.6289 \\
\bottomrule
\end{tabular}
\end{table*}

\subsection{RSSI Prediction}
\begin{figure}
    \centering
    \includegraphics[width=0.99\linewidth]{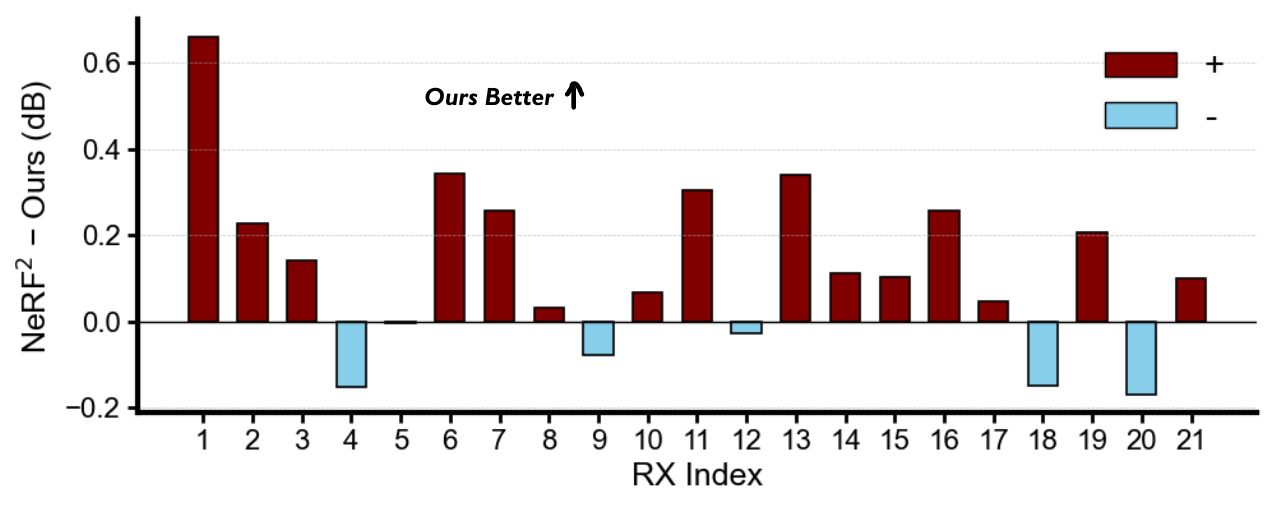}
    \caption{L1 loss (dB) difference between Ours and NeRF$^2$.}
    \label{fig:rssi_bar}
\end{figure}

\textit{Case setup: }We adapt the output head of SwiftWRF for RSSI prediction by applying a global average pooling layer to the generated spectrum map, effectively aggregating directional information into a single RSSI value. The spectrum resolution is set to 16×16, and the number of Gaussian primitives is reduced from 10k to 50 to ensure efficiency. For evaluation, we use a public BLE dataset \cite{Nerf2}, which contains RSSI measurements from multiple TXs at eight fixed RX locations. A separate SwiftWRF model is trained for each RX position over 100k iterations. As a baseline, we include the MRI method \cite{mri}, which estimates the RSSI at unsampled locations via interpolation based on a basic radio propagation model.\\
\textit{Result: }For overall comparison, our method achieves the lowest median l1 loss of \underline{\textbf{2.91}} dB, outperforming NeRF$^2$ at 3.05 dB and MRI at 6.74 dB. To further analyze performance at a finer granularity, we present a detailed per-RX comparison between SwiftWRF and NeRF$^2$ in Fig.~\ref{fig:rssi_bar}. The plot shows the difference in median l1 error for each RX index, where positive bars indicate cases where SwiftWRF performs better. SwiftWRF outperforms NeRF$^2$ on the majority of RX positions, with significant margins observed at RX 1, 6, 11, and 13.

\begin{table}[t]
\centering
\caption{Median L1 loss (dB) comparison across methods}
\begin{tabular}{lccc}
\toprule
\textbf{Method} & NeRF$^2$ & MRI & SwiftWRF \\
\midrule
\textbf{Median L1 loss (dB)↓} & 3.05 & 6.74 & \underline{\textbf{2.91}} \\
\bottomrule
\end{tabular}
\label{tab:l1_comparison}
\end{table}

\section{Conclusion}

In this paper, we present SwiftWRF, a novel framework that adapts GS for WRF modeling. Compared to NeRF-based methods, GS eliminates the need for volumetric sampling and expensive MLP queries along rays, enabling significantly faster and more scalable spectrum reconstruction. On top of this, our reformulation into a 2D Gaussian primitive system further reduces redundancy inherent to 3D representations in single-sided transceiver mobility settings. 

\textbf{Contribution: }SwiftWRF achieves high efficiency with reduced GPU memory consumption, enabling real-time spectrum synthesis at over 100k FPS and significantly faster training. Beyond spectrum reconstruction, it generalizes effectively to practical wireless tasks such as AoA and RSSI prediction, consistently outperforming existing baselines on both synthetic and real-world benchmarks. These results position GS as a promising new modeling paradigm for wireless propagation, effectively bridging the gap between geometry-driven simulations and data-driven field estimation.

\textbf{Discussion and Limitation: }
While SwiftWRF enables efficient spectrum reconstruction, our current model is trained offline and assumes a fixed environment during deployment. Incorporating lightweight online updates under transceiver mobility is a promising direction to further improve real-time channel estimation and system adaptability. Moreover, our current formulation models the measured spectrum outcomes rather than explicitly physical propagation effects such as visibility/path length or time-of-flight, and we keep the covariance updates fixed for stability. These design choices prioritize efficiency and stable optimization, but they may reduce physical interpretability. Although SwiftWRF achieves high fidelity in most evaluations, occasional synthesis failures and challenging transition regions suggest that further improvements in robustness and reconstruction quality are still needed. In addition, SwiftWRF does not currently support the NeRF$^2$ channel-estimation setting, which requires predicting high-dimensional CSI matrices from paired uplink/downlink measurements; developing a more stable multi-channel parameterization and training strategy is left for future work.


%



\ifCLASSOPTIONcompsoc
  \section*{Acknowledgments}
\else
  \section*{Acknowledgment}
\fi

This paper is supported by the National Natural Science Foundation of China (62371290, 62422111), National Key R\&D Program of China (2024YFB2907204), STCSM under Grant (24511107100), Shanghai Key Lab of Digital Media Processing and Transmission, Shanghai Jiao Tong University and Shanghai Municipal Education Commission (No.
2024AIYB002).

\ifCLASSOPTIONcaptionsoff
  \newpage
\fi

\bibliographystyle{IEEEtran}
\bibliography{refs}

@article{DeAlwis2021Survey,
  author       = {C. De Alwis and A. Kalla and Q.-V. Pham and P. Kumar and K. Dev and W.-J. Hwang and M. Liyanage},
  title        = {Survey on 6G frontiers: Trends, applications, requirements, technologies and future research},
  journal      = {IEEE Open Journal of the Communications Society},
  volume       = {2},
  pages        = {836--886},
  month        = feb,
  year         = {2021},
}

@article{schmitz2010efficient,
  title={Efficient rasterization for outdoor radio wave propagation},
  author={Schmitz, Arne and Rick, Tobias and Karolski, Thomas and Kuhlen, Torsten and Kobbelt, Leif},
  journal={IEEE Transactions on Visualization and Computer Graphics},
  volume={17},
  number={2},
  pages={159--170},
  year={2010},
  publisher={IEEE}
}

@article{3dgstvcg,
  title={3d gaussian splatting as new era: A survey},
  author={Fei, Ben and Xu, Jingyi and Zhang, Rui and Zhou, Qingyuan and Yang, Weidong and He, Ying},
  journal={IEEE Transactions on Visualization and Computer Graphics},
  year={2024},
  publisher={IEEE}
}

@article{zhang2024rf,
  title={RF-3DGS: Wireless Channel Modeling with Radio Radiance Field and 3D Gaussian Splatting},
  author={Zhang, Lihao and Sun, Haijian and Berweger, Samuel and Gentile, Camillo and Hu, Rose Qingyang},
  journal={arXiv preprint arXiv:2411.19420},
  year={2024}
}

@inproceedings{yang2025scalable,
  title={Scalable 3d gaussian splatting-based rf signal spatial propagation modeling},
  author={Yang, Kang and Du, Wan and Srivastava, Mani},
  booktitle={Proceedings of the 23rd ACM Conference on Embedded Networked Sensor Systems},
  pages={680--681},
  year={2025}
}

@article{raytracing,
  title={Ray tracing for radio propagation modeling: Principles and applications},
  author={Yun, Zhengqing and Iskander, Magdy F},
  journal={IEEE access},
  volume={3},
  pages={1089--1100},
  year={2015},
  publisher={IEEE}
}

@article{wrf,
  title={WRF-GS: Wireless Radiation Field Reconstruction with 3D Gaussian Splatting},
  author={Wen, Chaozheng and Tong, Jingwen and Hu, Yingdong and Lin, Zehong and Zhang, Jun},
  journal={arXiv preprint arXiv:2412.04832},
  year={2024}
}

@misc{vae,
  title={Auto-encoding variational bayes},
  author={Kingma, Diederik P and Welling, Max and others},
  year={2013},
  publisher={Banff, Canada}
}

@article{gan,
  title={Unsupervised representation learning with deep convolutional generative adversarial networks},
  author={Radford, Alec and Metz, Luke and Chintala, Soumith},
  journal={arXiv preprint arXiv:1511.06434},
  year={2015}
}

@inproceedings{Nerf2,
author = {Zhao, Xiaopeng and An, Zhenlin and Pan, Qingrui and Yang, Lei},
title = {NeRF2: Neural Radio-Frequency Radiance Fields},
year = {2023},
isbn = {9781450399906},
publisher = {Association for Computing Machinery},
address = {New York, NY, USA},
url = {https://doi.org/10.1145/3570361.3592527},
doi = {10.1145/3570361.3592527},
booktitle = {Proceedings of the 29th Annual International Conference on Mobile Computing and Networking},
articleno = {27},
numpages = {15},
location = {Madrid, Spain},
series = {ACM MobiCom '23}
}

@inproceedings{iArk,
author = {An, Zhenlin and Lin, Qiongzheng and Li, Ping and Yang, Lei},
title = {General-purpose deep tracking platform across protocols for the internet of things},
year = {2020},
isbn = {9781450379540},
publisher = {Association for Computing Machinery},
address = {New York, NY, USA},
url = {https://doi.org/10.1145/3386901.3389029},
doi = {10.1145/3386901.3389029},
booktitle = {Proceedings of the 18th International Conference on Mobile Systems, Applications, and Services},
pages = {94–106},
numpages = {13},
location = {Toronto, Ontario, Canada},
series = {MobiSys '20}
}

@misc{newrf,
      title={NeWRF: A Deep Learning Framework for Wireless Radiation Field Reconstruction and Channel Prediction}, 
      author={Haofan Lu and Christopher Vattheuer and Baharan Mirzasoleiman and Omid Abari},
      year={2024},
      eprint={2403.03241},
      archivePrefix={arXiv},
      primaryClass={cs.NI},
      url={https://arxiv.org/abs/2403.03241}, 
}

@manual{matlab2023a,
  title        = {{Indoor MIMO-OFDM Communication Link Using Ray Tracing}},
  author       = {{MATLAB}},
  organization = {MathWorks},
  year         = {2023},
  note         = {Available at: \url{https://www.mathworks.com/help/comm/ug/indoormimo-ofdm-communication-link-using-raytracing.html}},
}

@manual{matlab2023b,
  title        = {{Three-Dimensional Indoor Positioning with 802.11az Fingerprinting and Deep Learning}},
  author       = {{MATLAB}},
  organization = {MathWorks},
  year         = {2023},
  note         = {Available at: \url{https://www.mathworks.com/help/wlan/ug/threedimensional-indoor-positioning-with-802-11azfingerprinting-and-deep-learning.html}},
}

@ARTICLE{mri,
  author={Shin, Hyojeong and Chon, Yohan and Kim, Yungeun and Cha, Hojung},
  journal={IEEE Transactions on Mobile Computing}, 
  title={MRI: Model-Based Radio Interpolation for Indoor War-Walking}, 
  year={2015},
  volume={14},
  number={6},
  pages={1231-1244},
  doi={10.1109/TMC.2014.2345654}}

@article{he2018design,
  title={The design and applications of high-performance ray-tracing simulation platform for 5G and beyond wireless communications: A tutorial},
  author={He, Danping and Ai, Bo and Guan, Ke and Wang, Longhe and Zhong, Zhangdui and K{\"u}rner, Thomas},
  journal={IEEE communications surveys \& tutorials},
  volume={21},
  number={1},
  pages={10--27},
  year={2018},
  publisher={IEEE}
}

@article{mildenhall2021nerf,
  title={Nerf: Representing scenes as neural radiance fields for view synthesis},
  author={Mildenhall, Ben and Srinivasan, Pratul P and Tancik, Matthew and Barron, Jonathan T and Ramamoorthi, Ravi and Ng, Ren},
  journal={Communications of the ACM},
  volume={65},
  number={1},
  pages={99--106},
  year={2021},
  publisher={ACM New York, NY, USA}
}

@article{ye2025gsplat,
  title={gsplat: An open-source library for Gaussian splatting},
  author={Ye, Vickie and Li, Ruilong and Kerr, Justin and Turkulainen, Matias and Yi, Brent and Pan, Zhuoyang and Seiskari, Otto and Ye, Jianbo and Hu, Jeffrey and Tancik, Matthew and others},
  journal={Journal of Machine Learning Research},
  volume={26},
  number={34},
  pages={1--17},
  year={2025}
}

@article{tang2023dreamgaussian,
  title={Dreamgaussian: Generative gaussian splatting for efficient 3d content creation},
  author={Tang, Jiaxiang and Ren, Jiawei and Zhou, Hang and Liu, Ziwei and Zeng, Gang},
  journal={arXiv preprint arXiv:2309.16653},
  year={2023}
}

@inproceedings{huang20242d,
  title={2d gaussian splatting for geometrically accurate radiance fields},
  author={Huang, Binbin and Yu, Zehao and Chen, Anpei and Geiger, Andreas and Gao, Shenghua},
  booktitle={ACM SIGGRAPH 2024 conference papers},
  pages={1--11},
  year={2024}
}

@inproceedings{zhou2024drivinggaussian,
  title={Drivinggaussian: Composite gaussian splatting for surrounding dynamic autonomous driving scenes},
  author={Zhou, Xiaoyu and Lin, Zhiwei and Shan, Xiaojun and Wang, Yongtao and Sun, Deqing and Yang, Ming-Hsuan},
  booktitle={Proceedings of the IEEE/CVF conference on computer vision and pattern recognition},
  pages={21634--21643},
  year={2024}
}

@inproceedings{he2016deep,
  title={Deep residual learning for image recognition},
  author={He, Kaiming and Zhang, Xiangyu and Ren, Shaoqing and Sun, Jian},
  booktitle={Proceedings of the IEEE conference on computer vision and pattern recognition},
  pages={770--778},
  year={2016}
}

@inproceedings{chen2024rfcanvas,
  title={Rfcanvas: Modeling rf channel by fusing visual priors and few-shot rf measurements},
  author={Chen, Xingyu and Feng, Zihao and Sun, Ke and Qian, Kun and Zhang, Xinyu},
  booktitle={Proceedings of the 22nd ACM Conference on Embedded Networked Sensor Systems},
  pages={464--477},
  year={2024}
}

@article{li2025wideband,
  title={Wideband RF Radiance Field Modeling Using Frequency-embedded 3D Gaussian Splatting},
  author={Li, Zechen and Yang, Lanqing and Bian, Yiheng and Pan, Hao and Fu, Yongjian and Wang, Yezhou and Chen, Yi-Chao and Xue, Guangtao and Ren, Ju},
  journal={arXiv preprint arXiv:2505.20714},
  year={2025}
}

@inproceedings{wu20244d,
  title={4d gaussian splatting for real-time dynamic scene rendering},
  author={Wu, Guanjun and Yi, Taoran and Fang, Jiemin and Xie, Lingxi and Zhang, Xiaopeng and Wei, Wei and Liu, Wenyu and Tian, Qi and Wang, Xinggang},
  booktitle={Proceedings of the IEEE/CVF conference on computer vision and pattern recognition},
  pages={20310--20320},
  year={2024}
}

@article{kerbl20233d,
  title={3d gaussian splatting for real-time radiance field rendering.},
  author={Kerbl, Bernhard and Kopanas, Georgios and Leimk{\"u}hler, Thomas and Drettakis, George},
  journal={ACM Trans. Graph.},
  volume={42},
  number={4},
  pages={139--1},
  year={2023}
}

@article{sun2024splatter,
  title={Splatter a video: Video gaussian representation for versatile processing},
  author={Sun, Yang-Tian and Huang, Yihua and Ma, Lin and Lyu, Xiaoyang and Cao, Yan-Pei and Qi, Xiaojuan},
  journal={Advances in Neural Information Processing Systems},
  volume={37},
  pages={50401--50425},
  year={2024}
}

@inproceedings{zhang2024gaussianimage,
  title={Gaussianimage: 1000 fps image representation and compression by 2d gaussian splatting},
  author={Zhang, Xinjie and Ge, Xingtong and Xu, Tongda and He, Dailan and Wang, Yan and Qin, Hongwei and Lu, Guo and Geng, Jing and Zhang, Jun},
  booktitle={European Conference on Computer Vision},
  pages={327--345},
  year={2024},
  organization={Springer}
}

@inproceedings{yang2024deformable,
  title={Deformable 3d gaussians for high-fidelity monocular dynamic scene reconstruction},
  author={Yang, Ziyi and Gao, Xinyu and Zhou, Wen and Jiao, Shaohui and Zhang, Yuqing and Jin, Xiaogang},
  booktitle={Proceedings of the IEEE/CVF conference on computer vision and pattern recognition},
  pages={20331--20341},
  year={2024}
}

@article{liu2025d2gv,
  title={D2GV: Deformable 2D Gaussian Splatting for Video Representation in 400FPS},
  author={Liu, Mufan and Yang, Qi and Zhao, Miaoran and Huang, He and Yang, Le and Li, Zhu and Xu, Yiling},
  journal={arXiv preprint arXiv:2503.05600},
  year={2025}
}

@article{oestges2002deterministic,
  title={Deterministic channel modeling and performance simulation of microcellular wide-band communication systems},
  author={Oestges, Claude and Clerckx, Bruno and Raynaud, Lidwine and Vanhoenacker-Janvier, Danielle},
  journal={IEEE Transactions on Vehicular Technology},
  volume={51},
  number={6},
  pages={1422--1430},
  year={2002},
  publisher={IEEE}
}

@article{tong2018cooperative,
  title={Cooperative spectrum sensing: A blind and soft fusion detector},
  author={Tong, Jingwen and Jin, Ming and Guo, Qinghua and Li, Youming},
  journal={IEEE Transactions on Wireless Communications},
  volume={17},
  number={4},
  pages={2726--2737},
  year={2018},
  publisher={IEEE}
}

@inproceedings{orekondy2023winert,
  title={Winert: Towards neural ray tracing for wireless channel modelling and differentiable simulations},
  author={Orekondy, Tribhuvanesh and Kumar, Pratik and Kadambi, Shreya and Ye, Hao and Soriaga, Joseph and Behboodi, Arash},
  booktitle={The Eleventh International Conference on Learning Representations},
  year={2023}
}

@ARTICLE{rulebasedwrf1,
  author={Iskander, M.F. and Zhengqing Yun},
  journal={IEEE Transactions on Microwave Theory and Techniques}, 
  title={Propagation prediction models for wireless communication systems}, 
  year={2002},
  volume={50},
  number={3},
  pages={662-673},
  doi={10.1109/22.989951}}

@ARTICLE{rulebasedwrf2,
  author={Sarkar, T.K. and Zhong Ji and Kyungjung Kim and Medouri, A. and Salazar-Palma, M.},
  journal={IEEE Antennas and Propagation Magazine}, 
  title={A survey of various propagation models for mobile communication}, 
  year={2003},
  volume={45},
  number={3},
  pages={51-82},
  doi={10.1109/MAP.2003.1232163}}

@ARTICLE{6GIOTsurvey,
  author={Nguyen, Dinh C. and Ding, Ming and Pathirana, Pubudu N. and Seneviratne, Aruna and Li, Jun and Niyato, Dusit and Dobre, Octavia and Poor, H. Vincent},
  journal={IEEE Internet of Things Journal}, 
  title={6G Internet of Things: A Comprehensive Survey}, 
  year={2022},
  volume={9},
  number={1},
  pages={359-383},
  doi={10.1109/JIOT.2021.3103320}}

@article{yuan2025sed,
  title={SED-MVS: Segmentation-Driven and Edge-Aligned Deformation Multi-View Stereo with Depth Restoration and Occlusion Constraint},
  author={Yuan, Zhenlong and Yang, Zhidong and Cai, Yujun and Wu, Kuangxin and Liu, Mufan and Zhang, Dapeng and Jiang, Hao and Li, Zhaoxin and Wang, Zhaoqi},
  journal={IEEE Transactions on Circuits and Systems for Video Technology},
  year={2025},
  publisher={IEEE}
}

@ARTICLE{9598915,
  author={Alsabah, Muntadher and Naser, Marwah Abdulrazzaq and Mahmmod, Basheera M. and Abdulhussain, Sadiq H. and Eissa, Mohammad R. and Al-Baidhani, Ahmed and Noordin, Nor K. and Sait, Sadiq M. and Al-Utaibi, Khaled A. and Hashim, Fazirul},
  journal={IEEE Access}, 
  title={6G Wireless Communications Networks: A Comprehensive Survey}, 
  year={2021},
  volume={9},
  number={},
  pages={148191-148243},
  doi={10.1109/ACCESS.2021.3124812}}

@inproceedings{korhonen2012peak,
  title={Peak signal-to-noise ratio revisited: Is simple beautiful?},
  author={Korhonen, Jari and You, Junyong},
  booktitle={2012 Fourth international workshop on quality of multimedia experience},
  pages={37--38},
  year={2012},
  organization={IEEE}
}

@inproceedings{hore2010image,
  title={Image quality metrics: PSNR vs. SSIM},
  author={Hore, Alain and Ziou, Djemel},
  booktitle={2010 20th international conference on pattern recognition},
  pages={2366--2369},
  year={2010},
  organization={IEEE}
}

@INPROCEEDINGS{soft,
  author={Liu, Mufan and Chen, Jie and Wu, Gang and Ji, Lei and Wang, Hao},
  booktitle={GLOBECOM 2023 - 2023 IEEE Global Communications Conference}, 
  title={Soft-Ack based Outer Loop Link Adaptation for Latency-constrained 5G Video Conferencing}, 
  year={2023},
  volume={},
  number={},
  pages={388-393},
  doi={10.1109/GLOBECOM54140.2023.10437503}}

@article{liu2025channel,
  title={Channel Knowledge Maps for 6G Wireless Networks: Construction, Applications, and Future Challenges},
  author={Liu, Xingchen and Sun, Shu and Tao, Meixia and Kaushik, Aryan and Yan, Hangsong},
  journal={arXiv preprint arXiv:2505.24151},
  year={2025}
}

@article{zeng2021toward,
  title={Toward environment-aware 6G communications via channel knowledge map},
  author={Zeng, Yong and Xu, Xiaoli},
  journal={IEEE Wireless Communications},
  volume={28},
  number={3},
  pages={84--91},
  year={2021},
  publisher={IEEE}
}

@article{zeng2024tutorial,
  title={A tutorial on environment-aware communications via channel knowledge map for 6G},
  author={Zeng, Yong and Chen, Junting and Xu, Jie and Wu, Di and Xu, Xiaoli and Jin, Shi and Gao, Xiqi and Gesbert, David and Cui, Shuguang and Zhang, Rui},
  journal={IEEE communications surveys \& tutorials},
  volume={26},
  number={3},
  pages={1478--1519},
  year={2024},
  publisher={IEEE}
}

@inproceedings{li2022channel,
  title={Channel knowledge map for environment-aware communications: EM algorithm for map construction},
  author={Li, Kun and Li, Peiming and Zeng, Yong and Xu, Jie},
  booktitle={2022 IEEE Wireless Communications and Networking Conference (WCNC)},
  pages={1659--1664},
  year={2022},
  organization={IEEE}
}

@article{amballa2025can,
  title={Can NeRFs See without Cameras?},
  author={Amballa, Chaitanya and Basu, Sattwik and Wei, Yu-Lin and Yang, Zhijian and Ergezer, Mehmet and Choudhury, Romit Roy},
  journal={arXiv preprint arXiv:2505.22441},
  year={2025}
}

@article{nukapotula2025gsparc,
  title={GSpaRC: Gaussian Splatting for Real-time Reconstruction of RF Channels},
  author={Nukapotula, Bhavya Sai and Tripathi, Rishabh and Pregler, Seth and Kalathil, Dileep and Shakkottai, Srinivas and Rappaport, Theodore S},
  journal={arXiv preprint arXiv:2511.22793},
  year={2025}
}

\begin{IEEEbiography}
[{\includegraphics[width=1in,height=1.25in,clip,keepaspectratio]{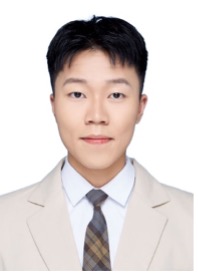}}]{Mufan Liu} received the B.Sc. degree in communication engineering from the University of Electronic Science and Technology of China, Chengdu, China, in 2023. He is currently pursuing the Ph.D. degree with the Cooperative MediaNet Innovation Center, Shanghai Jiao Tong University, Shanghai, China. His main research interests include adaptive streaming, joint source and channel coding and multimedia processing.
\end{IEEEbiography}

\begin{IEEEbiography}
[{\includegraphics[width=1in,height=1.25in, clip,keepaspectratio]{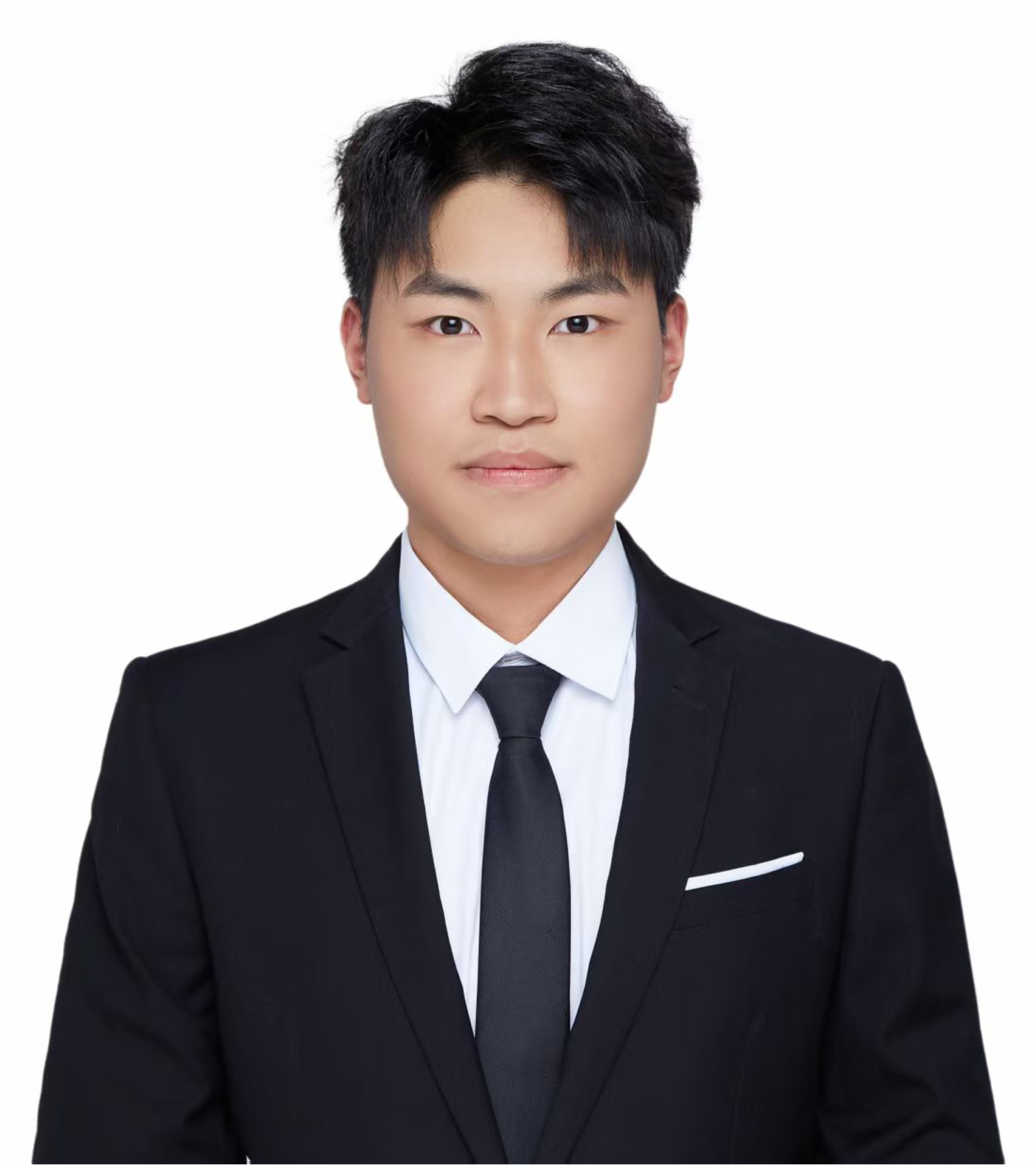}}]{Cixiao Zhang} received the B.Sc. degree in communication engineering from the University of Electronic Science and Technology of China, Chengdu, China, in 2023. He is currently pursuing the Ph.D. degree with the Cooperative MediaNet Innovation Center, Shanghai Jiao Tong University, Shanghai, China. His main research interests include Next-Generation Reconfigurable Antenna Systems, joint source and channel coding and multiple access technologies.
\end{IEEEbiography}

\begin{IEEEbiography}
[{\includegraphics[width=1in,height=1.25in,clip,keepaspectratio]{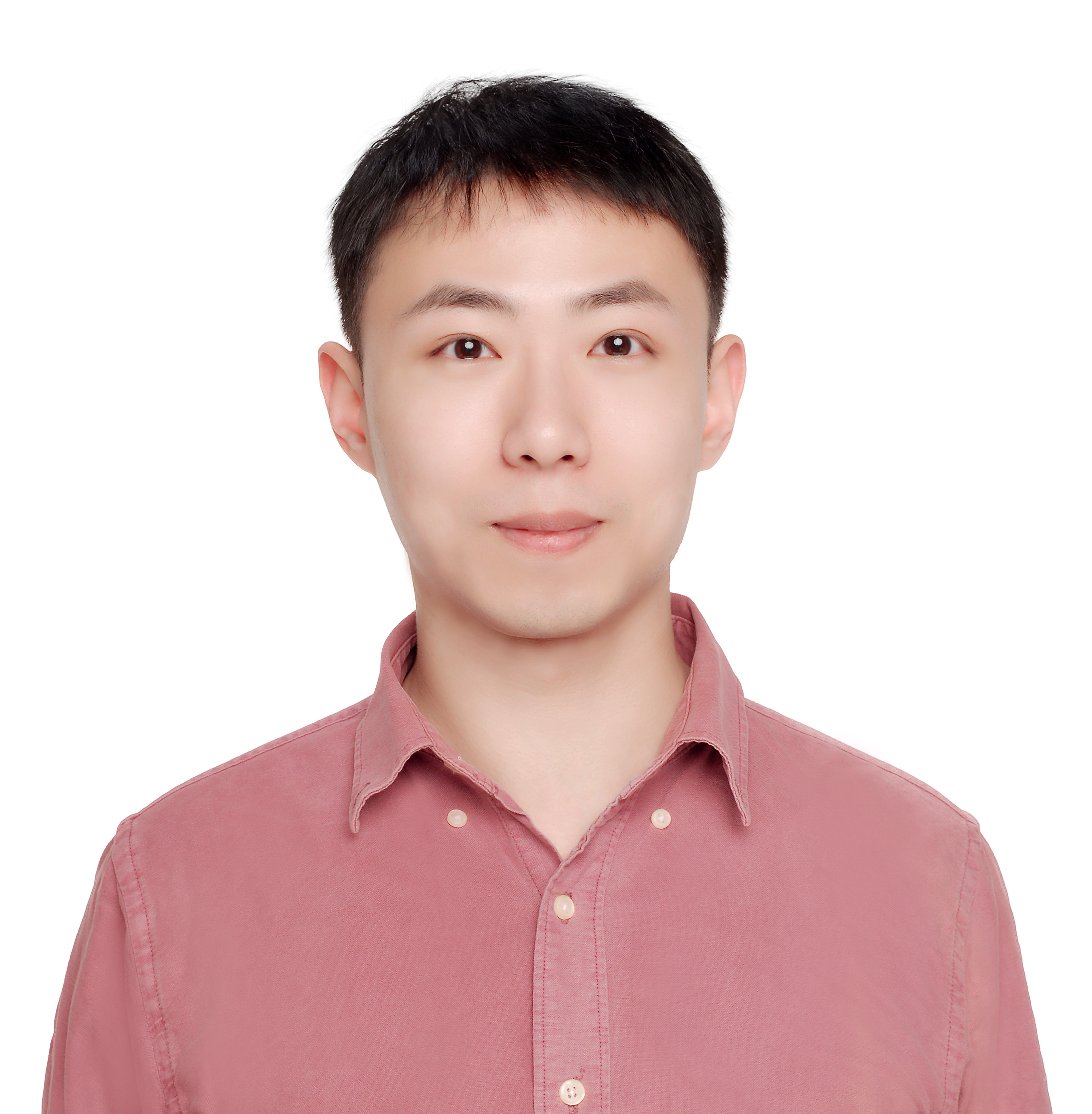}}]{Qi Yang} received the B.S. degree in communication engineering from Xidian University, Xi'an, China, in 2017, and Ph.D degree in information and communication engineering at Shanghai Jiao Tong University, Shanghai, China, 2022. He worked as a researcher in Tencent MediaLab from 2022 to 2024. Now, he joins University of Missouri–Kansas City as a research associate. He has published more than 35 conference and journal articles, including TPAMI, TIP, TVCG, TMM, TCSVT, ICML, CVPR, IJCAI, ACM MM, etc. He is also an active member in standard organizations, including MPEG, AOMedia, and AVS. His research interests include 3D/4D GS compact generation and compression, 3D point cloud and mesh quality assessment.
\end{IEEEbiography}

\begin{IEEEbiography}
[{\includegraphics[width=1in,height=1.25in,clip,keepaspectratio]{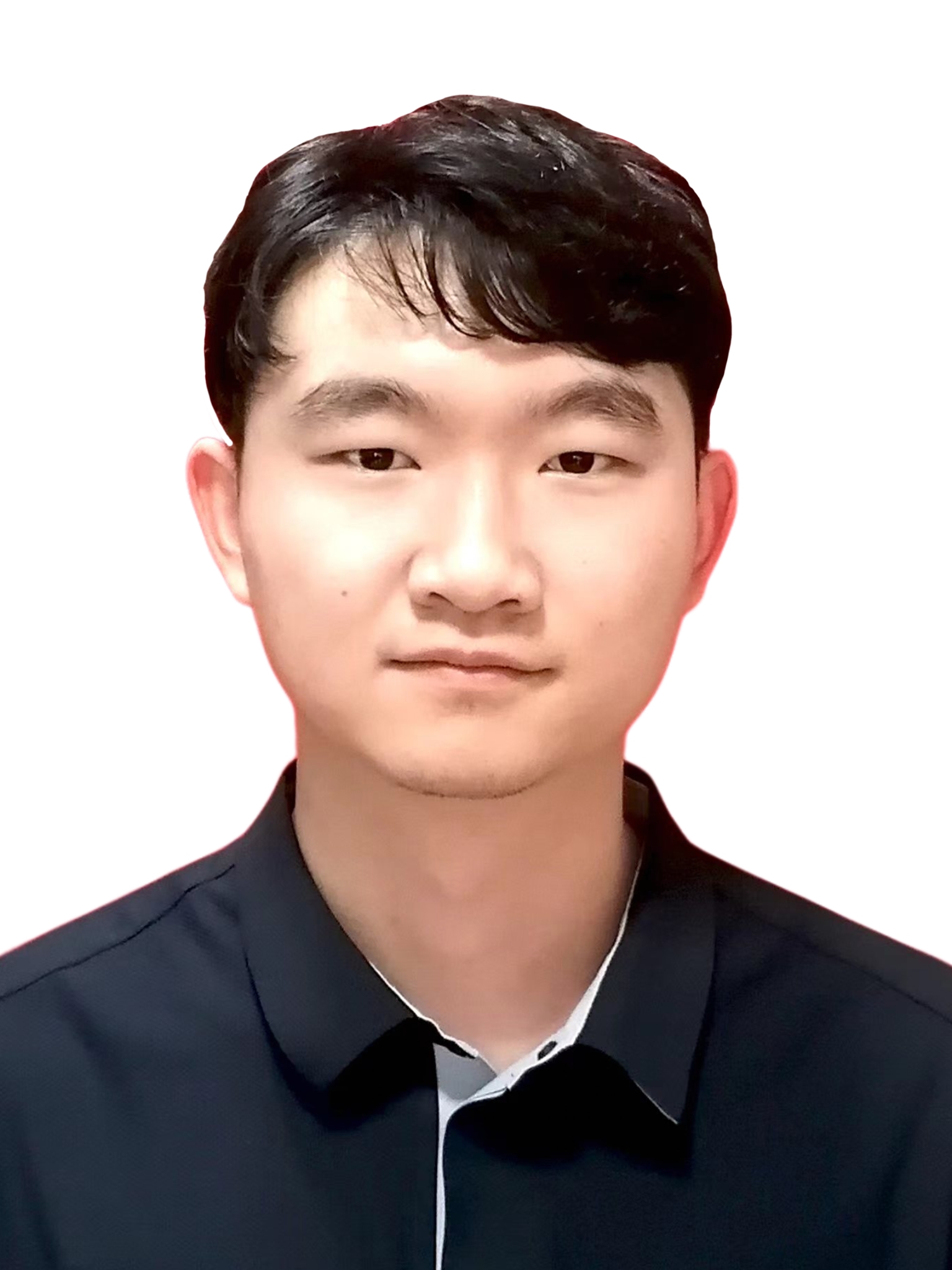}}]{Yujie Cao} received the B.Sc. degree in Information Engineering from Shanghai Jiao Tong University, Shanghai, China, in 2025. He is about to pursue the Ph.D. degree with the Cooperative MediaNet Innovation Center, Shanghai Jiao Tong University, Shanghai, China. His main research interests include semantic communications and 3DGS.
\end{IEEEbiography}

\begin{IEEEbiography}
[{\includegraphics[width=1in,height=1.25in,clip,keepaspectratio]{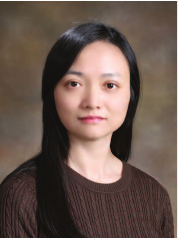}}]{Yiling Xu} received the B.S., M.S., and Ph.D degrees from the University of Electronic Science and Technology of China, in 1999, 2001, and 2004 respectively. From 2004 to 2013, she was a senior engineer with the Multimedia Communication Research Institute, Samsung Electronics Inc., South Korea. She joined Shanghai Jiao Tong University, where she is currently a professor in the areas of multimedia communication, 3D point cloud compression and assessment, system design, and network optimization. She is the associate editor of the IEEE Transactions on Broadcasting. She is also an active member in standard organizations, including MPEG, 3GPP, and AVS.
\end{IEEEbiography}

\begin{IEEEbiography}
[{\includegraphics[width=1in,height=1.25in,clip,keepaspectratio]{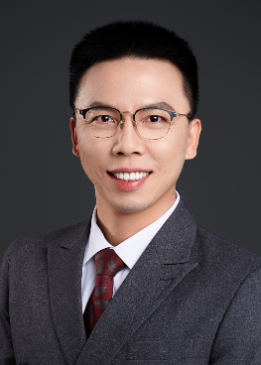}}]{Yin Xu} received his B.Sc. degree in Information Science and Engineering from Southeast University, China, in 2009. Subsequently, he obtained his M.Sc. and Ph.D. degrees in Electronics Engineering from Shanghai Jiao Tong University (SJTU) in 2011 and 2015, respectively. In 2011, he served as a visiting scholar at McGill University and INRS. Currently, he is an Associate Professor and Ph.D. Supervisor at SJTU. His research interests span key technologies such as channel coding and modulation, new waveforms, inter-tower communications, fluid/moveable antenna, and AI-assisted communications, along with the development of prototype systems for diverse communication systems including 5G/6G, satellite, broadcast, and short-range communications. He serves as an Associate Editor of Transactions on Broadcasting and is a recipient of the ‘6G Star Youth Scientist Award’. He also serves as co-chair, session chair, and keynote speaker at various major IEEE international conferences. Notably, he is actively involved in communication system standardization, serving as the chief 3GPP Standard delegate representing SJTU and holding the vice-chair position of Implementation Team 5 under the American Advanced Television Systems Committee (ATSC). His contributions to core physical layer technologies have been incorporated as key components into multiple globally reputable communication standards, including 3GPP 5G, ATSC 3.0, and SparkLink.
\end{IEEEbiography}

\begin{IEEEbiography}
[{\includegraphics[width=1in,height=1.25in,clip,keepaspectratio]{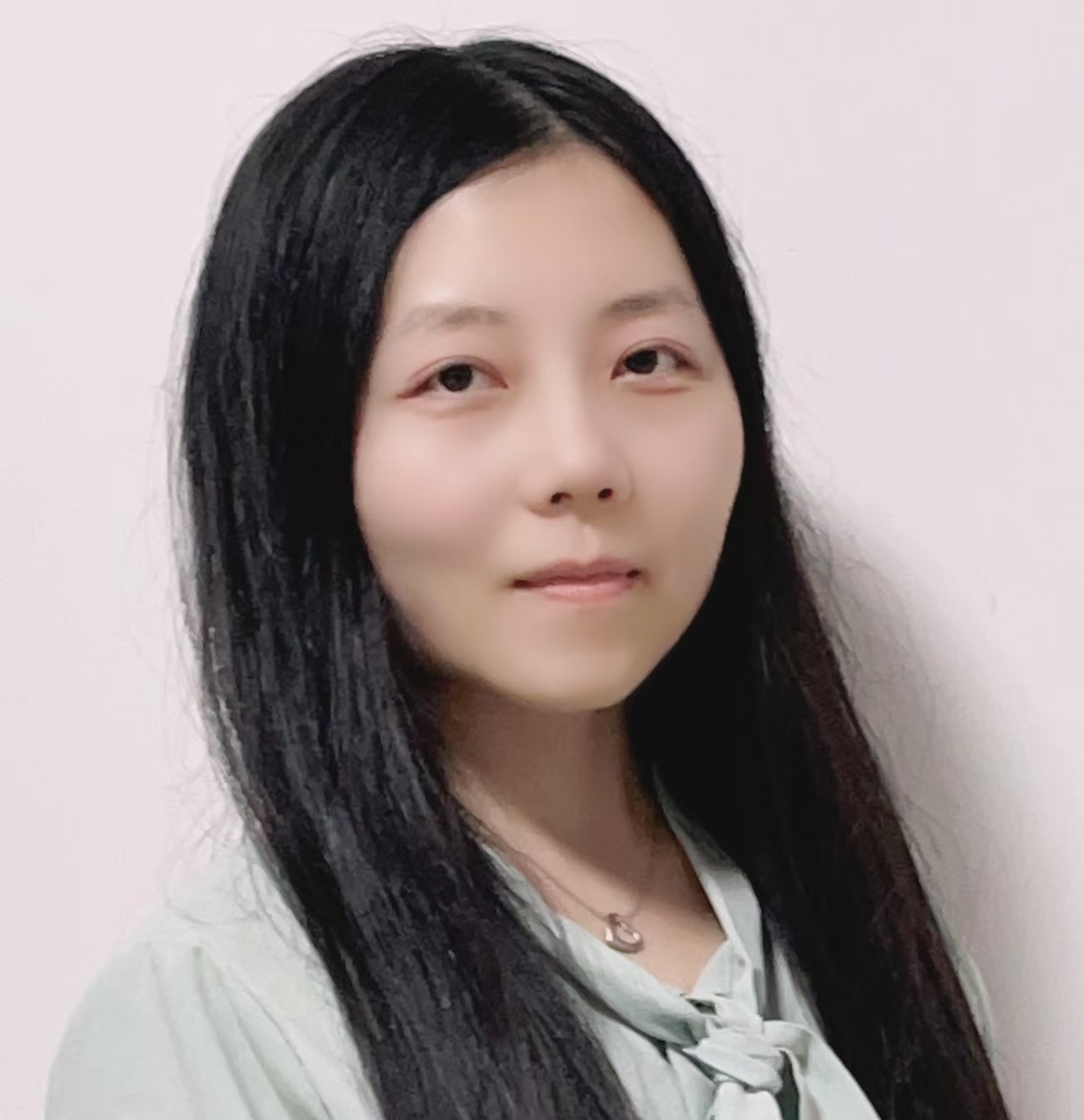}}]{Shu Sun} is a tenure-track associate professor in the School of Information Science and Electronic Engineering, Shanghai Jiao Tong University (SJTU), China. She obtained her B.S. degree in applied physics from SJTU in 2012 and Ph.D. degree in electrical engineering from New York University, USA, in 2018, and held summer internship positions at Nokia Bell Labs in 2014 and 2015. Shu also worked as a systems engineer at Intel Corporation. Her current research interests include channel modeling, millimeter-wave communications, integrated sensing and communication, and holographic MIMO. Dr. Sun received multiple international academic awards including the 2023 and 2017 IEEE Neil Shepherd Memorial Best Propagation Paper Awards, the 2017 Marconi Society Young Scholar Award, the IEEE VTC2016-Spring Best Paper Award, and the 2015 IEEE Donald G. Fink Award. She is an Associate Editor or Guest Editor of the \textsc{IEEE Transactions on Mobile Computing}, \textsc{IEEE Internet of Things Magazine}, and \textsc{IEEE Open Journal of Antennas and Propagation}. She also served as a Guest Editor for the \textsc{IEEE Journal on Selected Areas in Communications}.
\end{IEEEbiography}

\begin{IEEEbiography}
[{\includegraphics[width=1in,height=1.25in,clip,keepaspectratio]{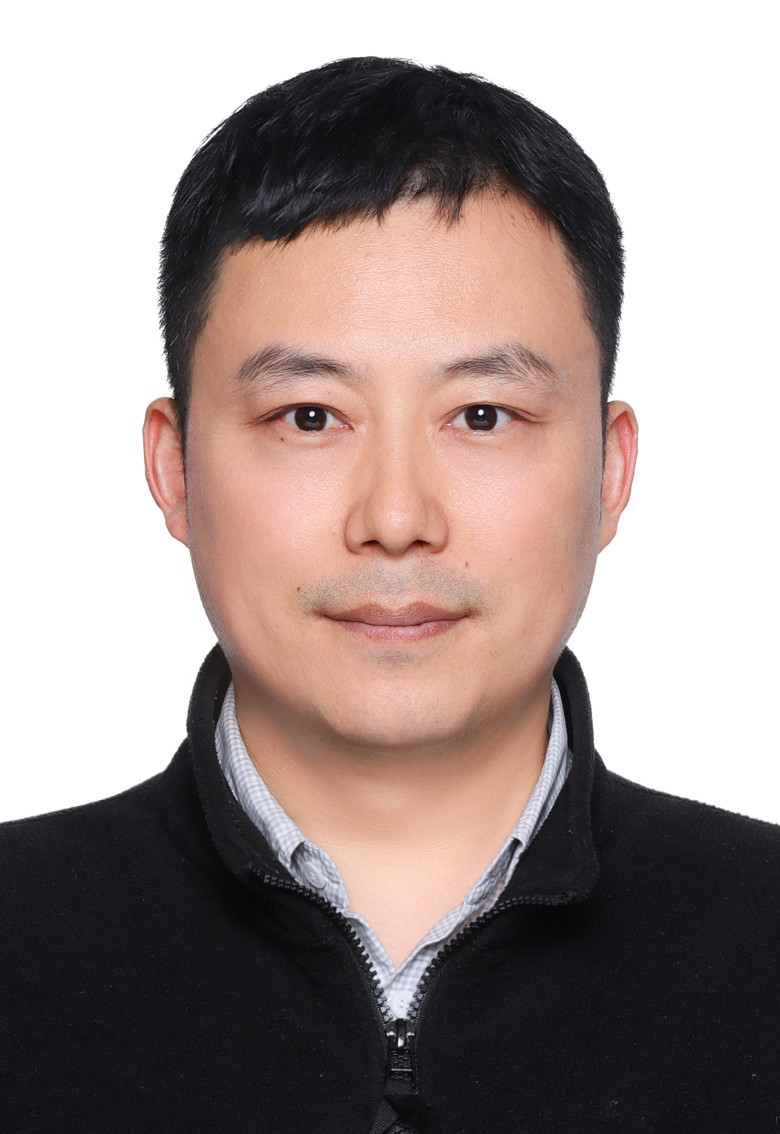}}]{Mingzeng Dai} obtained his Master’s degree in Telecommunication from Hangzhou Dianzi University in 2008. He is currently a Principal Researcher and 3GPP RAN2/3 delegate at the Wireless Research (WR) Lab of Lenovo Research. He has over 15 years’ experience in 3GPP standardization as a key contributor in many technical areas of 4G and 5G. He also serves as the co-leader of the RS03 Native AI and Cross-Domain AI initiatives within the O-RAN nGRG. His research interests primarily focus on 6G RAN architecture, Integrated Sensing and Communication (ISAC), and AI/ML-enabled RAN.
\end{IEEEbiography}

\begin{IEEEbiography}
[{\includegraphics[width=1in,height=1.25in,clip,keepaspectratio]{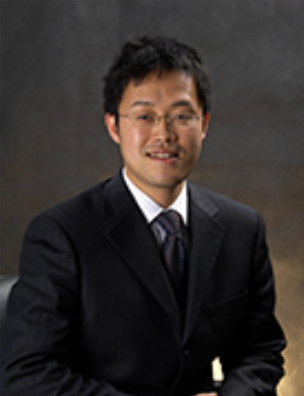}}]{Yunfeng Guan} received the Ph.D. degree from the Department of Electronics and Information Technology, Zhejiang University, Hangzhou, China, in 2003. Since 2003, he has been with the Institute of Wireless Communication Technology, Shanghai Jiao Tong University, where he is currently a Researcher with the Cooperative Medianet Innovation Center. His research interests include HDTV and wireless communications.
\end{IEEEbiography}

\end{document}